\documentclass{article}
\usepackage{arxiv}
\usepackage[T1]{fontenc}

\usepackage{amsthm}
\usepackage{graphicx}
\usepackage{subfig}
\usepackage{amsmath,amssymb,amsfonts}
\usepackage{mathrsfs}
\usepackage{xcolor}
\usepackage{textcomp}
\usepackage{manyfoot}
\usepackage[ruled,vlined]{algorithm2e}
\SetKwInOut{KwIn}{Input}
\SetKwInOut{KwOut}{Output}
\usepackage{bm}
\usepackage{listings}
\usepackage{tabularx}
\usepackage{booktabs}
\usepackage{multirow}
\usepackage{makecell}
\usepackage[
    colorlinks=true,
    linkcolor=blue,
    citecolor=blue!60!black,
    urlcolor=blue!50!black
]{hyperref}
\usepackage{siunitx}
\usepackage{colortbl}
\usepackage{caption}
\usepackage{lineno}
\usepackage{acronym}
\usepackage[utf8]{inputenc}
\usepackage[many]{tcolorbox}
\usepackage{tikz}
\usepackage{float}
\usepackage{natbib}

\usetikzlibrary{arrows.meta, positioning}
\newcolumntype{Y}{>{\centering\arraybackslash}X}

\DeclareMathOperator*{\argmin}{\arg\,\min}

\title{Robust-Multi-Task Gradient Boosting
\thanks{This manuscript is currently under review at *Expert Systems With Applications*.}}

\author{
  Seyedsaman Emami \\
  Escuela Polit\'ecnica Superior \\
  Universidad Aut\'onoma de Madrid \\
  Madrid\\
   \And
  Gonzalo Mart\'{\i}nez-Mu\~noz \\
  Escuela Polit\'ecnica Superior \\
  Universidad Aut\'onoma de Madrid \\
  Madrid\\
  \And
  Daniel Hern\'andez-Lobato \\
  Escuela Polit\'ecnica Superior \\
  Universidad Aut\'onoma de Madrid \\
  Madrid\\
}

\begin{document}
\maketitle
\begin{abstract}
      The objective of this study is to develop a robust
      boosting framework capable of handling
      heterogeneous and outlier tasks in Multi-Task Learning
      (MTL). Conventional MTL methods assume
      strong relatedness among tasks,
      which often fails in real-world scenarios
      involving adversarial or unaligned tasks that
      degrade performance. To address this
      limitation, we propose Robust
      Multi-Task Gradient Boosting (R-MTGB), a
      novel ensemble framework that explicitly
      models task heterogeneity within
      the gradient boosting paradigm.
      The methodology structures learning
      into three sequential stages: (1)
      shared representation learning to
      extract common patterns across tasks, (2)
      outlier-aware partitioning using a
      learnable task-specific parameter to
      separate and reweight outlier and non-outlier tasks,
      and (3) task-specific fine-tuning to
      refine individual predictors.
      Extensive experiments on both synthetic
      and real-world datasets demonstrate that
      R-MTGB consistently improves predictive accuracy,
      effectively identifies outlier tasks, and
      enhances generalization compared to
      state-of-the-art methods.
      The achieved results confirm that R-MTGB
      not only ensures robust performance and
      interpretability through task-level outlier
      scores but also provides a scalable and
      principled framework for reliable multi-task
      learning in heterogeneous environments.
\end{abstract}


\section*{List of Abbreviations}
\label{sec:abbreviations}

\begin{tcolorbox}[colframe=black!30!white, colback=white, coltitle=black!70!black, boxrule=0.2mm, rounded corners, width=0.9\textwidth, grow to right by=0.1\textwidth]
      \begin{acronym}[TLC]
            \acro{BTAMDL}{Boosting Tree-Assisted Multi-Task Deep Learning}
            \acro{CD}{Critical Distance}
            \acro{DNN}{Deep Neural Network}
            \acro{DP}{Data Pooling}
            \acro{DTR}{Decision Tree Regressor}
            \acro{FL}{Federated Learning}
            \acro{GB}{Gradient Boosting}
            \acro{MAE}{Mean Absolute Error}
            \acro{MDL}{Multi-Task Deep Learning}
            \acro{ML}{Machine Learning}
            \acro{MTGB}{Multi-Task Gradient Boosting}
            \acro{MTL}{Multi-Task Learning}
            \acro{R-MTGB}{Robust-Multi-Task Gradient Boosting}
            \acro{RMSE}{Root Mean Squared Error}
            \acro{ST}{Single-Task}
            \acro{TaF}{Task-as-Feature}
            \acro{TS}{Task-wise Split}
      \end{acronym}
\end{tcolorbox}

\section{Introduction}
\label{sec:intro}
\ac{ML} models are increasingly used in scenarios that require
learning multiple prediction tasks at the same time.
This approach, referred to as \ac{MTL},
involves learning multiple related or
unrelated tasks simultaneously by transferring knowledge
from one task to another~\citep{Zhang2022}.
The main objective of \ac{MTL} is to improve generalization
performance by utilizing task-specific information and leveraging
shared representations across tasks~\citep{Caruana1997}.
\ac{MTL} has demonstrated significant
potential in areas such as computer vision~\citep{Shen2024,Souček2024}
and healthcare~\citep{Liu2024,Tsai2025}.
By exploiting shared structures across tasks,
\ac{MTL} often achieves better generalization compared
to training separate models for each task.
However, practical applications frequently involve noisy,
diverse, or even adversarial task environments,
making robustness an essential consideration.
In such cases, conventional \ac{MTL} models can
experience substantial performance degradation when some
tasks are corrupted, poorly defined,
or entirely unrelated to the other tasks~\citep{Yu2007}.

Beyond \ac{MTL}, \ac{GB} variants
have become one of the most effective techniques
in supervised learning, especially when applied with
decision trees~\citep{Maciej2016,Chen2016,Bentejac2021,Ravid2022}.
Specifically, \ac{GB} fits a function that explains the target
values associated to each input. This function is obtained by
combining several predictors, each one obtained by performing a
gradient step in function space minimizing a
particular loss function~\citep{Friedman2001}.
Building on this success, Multi-Task \ac{GB} extends the
\ac{GB} framework to handle multiple related learning tasks
simultaneously. For this, a function is fit for each task.
Importantly, however,
such a function is obtained as the sum of two functions.
Namely, a common function that captures the shared structure among tasks,
and a task-specific function
that
accounts for task-specific
deviations~\citep{Olivier2011,Emami2023}.
This formulation enables implicit data
sharing and acts as a regularizer,
improving generalization—particularly
when tasks are similar but not identical.
Unlike \ac{ST} learning, which ignores potential
synergies between tasks, or \ac{DP}, which treats
all tasks as identical, multi-task boosting leverages
shared structure while respecting task heterogeneity.
Empirical results have shown that this method consistently
outperforms standard boosting approaches in scenarios
where tasks are moderately related~\citep{zhang2012mtboost,
      Bellot2018,Emami2023}.

Critically, the multi-task variants of \ac{GB},
such as \ac{MTGB}, rely on the key assumption of a
shared function across all tasks~\citep{Emami2023}.
This need not be the case when some of the tasks are outlier tasks,
\emph{i.e.}, they are tasks that do not share any relation with the
other tasks. Outlier tasks simply differ significantly from the
other tasks and may deteriorate the \ac{MTL} process.  As a matter
of fact, in real-world \ac{MTL} scenarios, tasks often exhibit
significant heterogeneity~\citep{Yu2007,Gong2012},
reducing the effectiveness of standard \ac{MTL}
approaches~\citep{Yu2007}. Under such conditions,
the performance of methods like \ac{MTGB} can be
severely impaired.  Therefore, robustness to outlier
tasks and to variations in task difficulty or data quality
becomes a critical feature of \ac{MTL} methods. Notwithstanding,
robust techniques for \ac{MTL} based on \ac{GB} remain largely unexplored.

In this paper, we propose a novel multi-task boosting algorithm called
\ac{R-MTGB},
that can learn across tasks with varying
degrees of relatedness. \ac{R-MTGB} introduces a structured ensemble
learning framework composed of three sequential blocks.
In the first block, the model learns a global shared representation
that captures commonalities across all tasks. The second block
distinguishes between outlier and non-outlier tasks by optimizing
a regularized task-specific parameter jointly. This enables
adaptive weighting of task contributions
and mitigates the influence of outlier tasks on
the shared function. Finally, the third block performs
fine-tuning by learning task-specific predictors, enabling
the model to capture the nuances of individual tasks.
Importantly, however, the level of contribution of each block
to the overall learning process can be adjusted to the observed data.
For this, one simply has to change the number of boosting predictors used
in that particular block. To be specific, the number of
predictors used in each block is a hyperparameter that can be tuned,
e.g, using a cross-validation grid search.
This modular design allows \ac{R-MTGB} to dynamically balance
shared-learning
and task-specific adaptation, improving generalization across
heterogeneous task
sets.  As a result, the model is robust to outlier tasks.
Besides this, it is also scalable to large datasets, and adaptable to
a wide range
of loss functions. Finally, \ac{R-MTGB} not only improves
robustness and generalization performance,
but also enhances interpretability by allowing the clustering
of tasks into non-outlier
and outlier categories based on the results of the learning process.

The key contributions of this study are as follows:
\begin{itemize}
      \item \textbf{A novel integration of outlier task detection
                  into gradient boosting for multi-task learning:}
            Unlike previous boosting-based \ac{MTL}
            methods, task-level outlier-aware partitioning
            is incorporated directly into the boosting
            iterations, enabling adaptive emphasis on
            informative tasks during training.

      \item \textbf{A principled three-stage design
                  developed for heterogeneous task environments:}
            A sequential architecture comprising shared
            representation learning, outlier-aware task partitioning,
            and task-specific refinement, is introduced,
            theoretically motivated, and empirically validated.
            This design balances generalization through
            shared patterns and specialization
            via per-task refinement, while remaining robust to outlier tasks.

      \item \textbf{A unified boosting formulation
                  that generalizes existing models:}
            It is shown that several established approaches
            emerge as special cases of the proposed model
            when certain components are omitted,
            highlighting its role as a flexible
            generalization rather than a simple
            combination of existing methods.

      \item \textbf{Extensive empirical validation with
                  robustness analysis:}
            The proposed approach improves predictive
            accuracy and produces interpretable task-level
            outlier scores across synthetic and
            real-world benchmarks.

\end{itemize}

Collectively, these innovations enable \ac{R-MTGB} to
bridge the gap arising from the lack of robust multi-task
boosting methods capable of handling heterogeneous and
outlier tasks. Traditional multi-task boosting frameworks,
such as \ac{MTGB}, assume a uniform degree of
relatedness among tasks and thus become vulnerable when
confronted with unaligned tasks. \ac{R-MTGB}
overcomes this limitation through an automatic
outlier detection mechanism that assigns extreme,
opposite weights to outlier and non-outlier tasks.
This mechanism allows the model to automatically
separate and adapt to heterogeneous task behaviors,
reducing the disruptive influence of anomalous tasks while
preserving the shared structure among related ones.
Consequently, \ac{R-MTGB} provides a unified boosting
framework that achieves robustness and interpretability
without sacrificing predictive accuracy.

The remainder of this paper is structured as follows.
Section~\ref{sec:related_work} reviews prior studies on
\ac{MTL}, \ac{GB}, and multi-task boosting frameworks,
highlighting their limitations and comparing
them with the proposed approach.
Section~\ref{sec:method} introduces the developed
methodology, including its mathematical foundations
and theoretical analysis.
Section~\ref{sec:experiments} presents the experiments
conducted on both synthetic and real-world datasets, along with
a detailed discussion of the results.
Finally, Section~\ref{sec:conclusions} concludes
the study with a summary of the key findings.

\section{Related Work}
\label{sec:related_work}
This section reviews prior work relevant to the
proposed approach. We begin by
introducing the core ideas and categories of
\ac{MTL} in Subsection~\ref{sub_sec:mtl_related_work}.
In Subsection~\ref{sub_sec:gb_related_work}, we
summarize the development of \ac{GB} and its variants.
Finally, in Subsection~\ref{sub_sec:gb_mtl},
we discuss prior attempts to apply boosting methods to
\ac{MTL} problems and highlight the differences
between these approaches and ours.

\subsection{Multi-Task Learning}
\label{sub_sec:mtl_related_work}
\ac{MTL} is an \ac{ML} approach in which multiple tasks
are learned simultaneously, allowing shared knowledge across
functions to improve overall
performance~\citep{Caruana1997}.
The core assumption in \ac{MTL} is
that tasks within a given dataset
are related~\citep{Caruana1997,li2015multi}. By
leveraging transfer learning, \ac{MTL} enables models to
use information gained from one task to enhance learning and
generalization on related tasks, leading to more robust and
adaptable systems compared to training separate
models for each task~\citep{Zhang2022}.

Several well-defined approaches to exploring \ac{MTL}
have been studied, including feature learning, low-rank
parameterization, task clustering,
task relationship modeling, and decomposition methods~\citep{Zhang2022}.
In feature learning, the objective is to discover a
shared representation across multiple tasks by leveraging
shared features. This approach has been implemented in
various \ac{ML} models,
including neural
networks~\citep{Caruana1997,Liao2005}
and deep neural
networks~\citep{Zhang2014,LI2014,Liu2015,Zhang2015}.

The Low-Rank methodology, on the other hand, is
designed to capture the relatedness among
tasks by assuming that the parameter matrix
across tasks lies in a low-rank subspace
\citep{ando2005framework}.
This implies the existence of shared latent
factors among tasks. The objective is to
minimize a joint loss function over the weight
matrix, subject to a low-rank
constraint (often via nuclear norm
regularization or matrix factorization).
Recent studies have applied this approach
to develop \ac{MTL} approaches in various areas,
such as improving parameter-efficient
training of multi-task models
\citep{Agiza2024}, identifying outliers~\citep{Chen2011},
and reconstructing low-rank weight matrices~\citep{Han2016}.

Another approach in \ac{MTL} involves grouping
related tasks into clusters,
as first proposed by~\cite{Thrun1996}, where it exploits the
shared structure within each cluster to
enhance learning. Later, a theoretical framework for
\ac{MTL} based on clustering tasks and assigning each cluster
to one of a limited number of shared hypotheses was
proposed~\citep{Crammer2012}.
This hard-assignment strategy facilitates learning from
limited data while effectively controlling model complexity.

Another category in \ac{MTL} focuses on approaches that
encourage the model to treat the average of task-specific
parameters as a central assumption,
based on the idea that tasks are inherently
similar~\citep{Evgeniou2004,Parameswaran2010}.
Other studies regularize the objective function by
measuring pairwise task similarities~\citep{Evgeniou2005}
or controlling task relatedness~\citep{Kato2007}.

Lastly, the decomposition approach involves breaking
down model parameters into shared and task-specific components.
This enables the model to learn shared patterns across tasks
while also capturing nuances unique to individual tasks.
A study by~\cite{Han2015} introduced an approach that
simultaneously learns
both shared and task-specific parameters directly from data by
implementing a layered decomposition of the parameter matrix,
with each layer representing a level in the task hierarchy.
Another study decomposes the model parameters for each task into
shared components and task-specific deviations,
applying a methodology to learn multiple related
parameters tasks simultaneously~\citep{Evgeniou2004}.
This approach allows for better control over the shared
information, with each task parameter vector represented
as the sum of a shared vector and a task-specific offset.
In a related line of work, but using a different \ac{ML} model,
\ac{MTGB} was introduced~\citep{Emami2023}. This approach
explicitly incorporates both shared and task-specific components
through a two-phase process. In the first phase,
a common set of models is trained to capture patterns
shared across all tasks. In the second phase,
separate models are added for each task to learn the
pseudo-residual information specific to that task.

Recent advancements in \ac{MTL}
have been primarily fueled by deep learning frameworks
capable of capturing hierarchical and shared
representation~\citep{Zhang2022}.
These approaches can generally be classified into three
primary categories. The first category focuses on
learning a unified feature representation across multiple
tasks by sharing the initial layers of the
network~\citep{Zhang2014,Liu2015,Zhang2015,LI2014}.
The second category employs adversarial learning techniques
to obtain a common representation suitable for multiple
tasks~\citep{Shinohara2016}.
The third category, represented by the cross-stitch network,
aims to learn distinct yet interrelated feature representations
for different tasks~\citep{Misra2016}.

Despite the successes of deep learning-based \ac{MTL} methods
in domains such as computer vision~\citep{Vandenhende2022,Fontana2024}
and natural language processing~\citep{Liu2018,Chen2024},
these architectures inherit some of the same limitations when
applied to tabular data. In particular,
multiple recent studies show that neural-network
approaches often underperform tree-ensemble or
boosting methods on standard tabular classification and
regression benchmarks, and require heavier hyperparameter
tuning, more training effort,
and less interpretability than ensemble-based
methods~\citep{Grinsztajn2022,McElfresh2023,Borisov2024}.
These limitations carry over into \ac{MTL} settings:
when tasks involve outlier tabular data, a deep \ac{MTL}
model may struggle to extract the correct inductive biases
or to adapt to outlier tasks, potentially resulting in
sub-optimal performance~\citep{Aoki2022,Aakarsh2023}.
Hence, \ac{MTL}
frameworks based purely on deep networks may be
less optimal in tabular data environments compared to
ensemble-based methods such as boosting
ensembles~\citep{Zhang2019,Jiang2020Boosting}.

\subsection{Gradient Boosting}
\label{sub_sec:gb_related_work}
In the context of tabular datasets and supervised
learning models, ensemble learning has demonstrated strong
performance in solving a wide range of
\ac{ML} problems~\citep{Ravid2022},
including classification and
regression~\citep{Maciej2016,Lakshminarayanan2017,
      Hongliang2018,Xia_2023_ICCV,Aybike2024}.
Ensemble models work
by combining multiple base learners to construct a
more robust and accurate final model~\citep{zhou2012ensemble,Azal2024}.

\ac{GB} is among the most successful ensemble methods.
It builds predictive models by
sequentially adding base learners (typically \ac{DTR}) to correct the
residuals of preceding models~\citep{Friedman2001}.
This results in a learning process that minimizes a loss function
by performing gradient descent in functional space.
More precisely, adding each new predictor can be seen as a step in functional
space with the goal of minimizing the loss function (\emph{i.e.}, the squared error for
regression or the cross-entropy for classification).  The result of the
learning process is a function from inputs to targets
that is optimal according to the particular loss function employed.

A faster variant of \ac{GB} is XGBoost, which introduces regularization into
the objective function and employs an improved branch-splitting method
in \ac{DTR}, resulting in faster training and enhanced
accuracy~\citep{Chen2016}.
Another notable advancement is
LightGBM, which accelerates \ac{GB} through novel
sampling strategies and feature bundling methods,
achieving high speed and performance~\citep{Ke2017}.
CatBoost, further addresses prediction shifts in \ac{GB}
by introducing a
permutation-based technique~\citep{Prokhorenkova2018}.

The strong performance of \ac{GB}
has led to the development of multi-class classification and
multi-output regression models. These models
extend \ac{GB} and its variants
to address such problems more efficiently by restructuring \ac{GB}
variants to support multi-output and multi-class problems within their
loss functions and base learners.
Gradient-Boosted Decision Trees for Multiple Outputs
represents the multi-output extension
of XGBoost~\citep{Zhang2021},
while Condensed-Gradient Boosting is a multi-output version
of \ac{GB} that employs multi-output
\ac{DTR}~\citep{Emami2025}, which is also less
complex than previous \ac{GB} variants in
terms of both time and space requirements.
These developments highlight flexibility and potential
of \ac{GB} framework for tackling more complex supervised
learning problems involving multiple tasks.

\subsection{Boosting Multi-Task Learning}
\label{sub_sec:gb_mtl}

The first study to propose a multi-task
boosting approach is~\cite{Olivier2011}, where
the authors employed \ac{GB} to design a customized
\ac{MTL} framework.
Their method maintains \( T + 1 \) base learners composed
of (\ac{DTR}s): one global model to capture shared structure among
all tasks and \( T \) task-specific models to account for
individual task nuances. At each boosting iteration, the
algorithm adds a new base learner to either the global
or a task-specific model, depending on which yields the
greatest reduction in the overall objective,
determined via steepest descent. \( \ell_1 \)-norm
regularization is employed to promote sparsity in the
learned functions. The overall objective is to
iteratively optimize a shared loss function across tasks
by selecting the direction (i.e., base learner and task)
that most improves the loss, as approximated through a
first-order Taylor expansion.

Another boosting-based \ac{MTL} framework is \ac{BTAMDL},
which integrates \ac{DTR} with \ac{DNN}~\citep{Jiang2020}.
The boosting process is implemented in two distinct stages.
In the first stage, a \ac{MDL} network is
trained on several related tasks to learn
shared representations, thereby leveraging
data-rich tasks to support those with limited data.
In the second stage, the output of the final hidden
layer of the \ac{MDL} network is used
as input features for training a \ac{GB} model.

Another study similar to the one proposed
in~\cite{Olivier2011} is
\ac{MTGB}. However, \ac{MTGB} differs in structure,
framework, and optimization strategy~\citep{Emami2023}.
Specifically, \ac{MTGB} builds on \ac{GB} by explicitly separating the
learning process into two stages. First, learning a shared
function for each task by fitting shared base learners across tasks.
Second, learning a task-specific function for each task by fitting task-specific base learners.
The final function used to predict targets given inputs for each task
is a combination of the two aforementioned functions.
In summary, the optimization (carried out using gradient
descent in the functional space) is performed in two phases:
first, a shared loss function is
minimized using the combined data from all tasks;
then, task-specific loss functions are
optimized separately for each task.
A limitation of this work is that outlier tasks, which
significantly differ from the other tasks by sharing no information at
all with them, may deteriorate the process of fitting the
shared function described above.

An alternative approach that differs
significantly from previous studies
is Boosted-\ac{MTL}
framework, which is based on a
\ac{FL} paradigm~\citep{Liu_Haizhou2024}.
This framework operates in two sequential stages.
First, in the \textit{global learning} stage,
multiple districts collaborate through a
privacy-preserving federated \ac{GB} scheme,
known as FederBoost, to learn shared load
patterns. Second, in the \textit{local learning} stage,
each district independently fine-tunes a local model to capture
its district-specific load characteristics.
The final model is constructed as the sum of the
global and local \ac{GB} models.

In contrast to restructuring the \ac{GB} framework,
\ac{TS}-\ac{GB} introduced a task-specific
splitting mechanism for \ac{DTR}~\citep{Mingcheng2021},
replacing the standard criterion with one based
on task-specific performance, termed task gain.
A split is performed only when the
negative impact on other tasks does not exceeds a predefined threshold.
Later, an extension of \ac{TS}-\ac{GB} was introduced to
address the issue of imbalanced data by
proposing two approaches: {$\text{\ac{TS}-\ac{GB}}_{\beta}$},
which revises the task gain ratio to be
more sensitive to the number of affected tasks
rather than the number of instances;
and {$\text{\ac{TS}-\ac{GB}}_{\kappa}$}, which reweights datasets
using a softmax-based method to balance data
distribution across tasks~\citep{Handong2022}.
These adaptations enhance both overall and
task-specific prediction performance without compromising
the accuracy for minority labels,
as reported in a recent preprint~\citep{ZhenZhe2022}.

\subsection{Comparative analysis}

Our proposed method falls into the first category of
boosting-based \ac{MTL} approaches. However, unlike tree-based models or
the \ac{FL} framework, our method redefines the
ensemble structure to directly address challenges specific
to \ac{MTL}. Importantly, while prior work in this category
has largely focused on modeling shared and task-specific
patterns, they have not addressed the presence of
outlier or adversarial tasks, which can degrade overall
model performance. Table~\ref{tab:mtl_boosting_comparison} summarizes key design
choices across representative boosting-based \ac{MTL} methods.
Specifically, we compare whether a method explicitly
(i) learns a \emph{shared representation}
across tasks, (ii) models \emph{task-specific}
components, and (iii) handles \emph{outlier/adversarial tasks}
\emph{within the GB fitting process}.
The table shows that the proposed approach, \ac{R-MTGB}, is
the only one fulfilling the three considered criteria.

\begin{table*}[t]
      \caption{Comparison of boosting-based \ac{MTL}.
            (Symbols: $\checkmark$ = present; $\times$ =
            absent).}
      \centering
      \resizebox{\textwidth}{!}{%
            \begin{tabular}{lcccc}
                  \toprule
                  \textbf{Method}                       & \textbf{Shared Rep.}       & \textbf{Task-specific Modeling} & \textbf{Outlier Task Handling } \\
                  \midrule
                  Boosted \ac{MTL}~\citep{Olivier2011}  & $\checkmark$               & $\checkmark$                    & $\times$                        \\
                  \ac{MTGB}~\citep{Emami2023}           & $\checkmark$               & $\checkmark$                    & $\times$                        \\
                  \ac{BTAMDL}~\citep{Jiang2020}         & $\checkmark$               & $\checkmark$                    & $\times$                        \\
                  FederBoost~\citep{Liu_Haizhou2024}    & $\checkmark$ (global)      & $\checkmark$ (local)            & $\times$                        \\
                  \ac{TS}-\ac{GB}~\citep{Mingcheng2021} & $\checkmark$ (shared tree) & $\checkmark$ (splits adapted)   & $\times$                        \\
                  \midrule
                  \ac{R-MTGB} (Ours)                    & $\checkmark$               & $\checkmark$                    & $\checkmark$                    \\
                  \bottomrule
            \end{tabular}
      }
      \label{tab:mtl_boosting_comparison}
\end{table*}

Existing boosting-based \ac{MTL} methods primarily
decompose the predictor into a shared part and per-task components,
and then decide where to add base learners
(e.g., global vs.\ task-specific in \cite{Olivier2011}
or \cite{Emami2023}). \ac{BTAMDL}~\citep{Jiang2020} introduces an
\ac{MDL}-based regularization to balance global and
task-specific learners. This encourages careful
information sharing, but does not explicitly identify adversarial or outlier tasks.
\ac{TS}-\ac{GB}~\citep{Mingcheng2021} addresses the
negative task transfer problem by constraining tree splits at
the node level, thereby reducing harmful feature-wise splits.
Nevertheless, it does not \emph{learn} a notion of task outlierness
nor it reweights task contributions during \ac{GB} training.
Therefore, its robustness is heuristic rather
than adversarial-task driven. \ac{FL} variants such
as FederBoost~\citep{Liu_Haizhou2024} combine global
collaboration (distributional via \ac{FL}) with local
refinement, improving privacy and distributional robustness. Yet, they also
do not \emph{jointly optimize} any task-outlier variable that influence each
base learner contributions inside boosting.
This reveals a clear research gap in the current literature:
while existing multi-task boosting methods can model shared
and task-specific information, none can automatically identify
and adapt to outlier or adversarial tasks within the
boosting process.
By contrast, our proposed method, \ac{R-MTGB},
incorporates a learnable mechanism, parameterized within the boosting loop,
which enables a soft partition of tasks into outlier and non-outlier components.
Thus, \ac{R-MTGB} provides principled robustness against adversarial
or outlier tasks and does not rely on heuristic adjustments.
Consequently, outlier tasks are down-weighted where
they would corrupt the shared structure among related tasks,
yet, each task still benefits from task-specific fine-tuning.
To our knowledge, no prior \ac{GB}-based \ac{MTL} approach
learns such an \emph{task-specific parameter}
that (i) detects outlier tasks \emph{during}
boosting and (ii) adapts their learning process to a
separate component, thereby providing principled
robustness to task heterogeneity. This not only
improves robustness and generalization performance, but also enhances
interpretability by allowing the clustering of tasks into non-outlier
and outlier categories based on the results of the learning process.

\section{Methodology}
\label{sec:method}
This section details the methodology used in this study.
It begins with the preliminaries and notation
(Subsection~\ref{sub_sec:preliminaries}),
followed by an introduction to \ac{MTL}
(Subsection~\ref{sub_sec:mt}). Next, an overview
of \ac{GB} framework is provided
(Subsection~\ref{sub_sec:gb}), leading to the
presentation of the proposed \ac{R-MTGB}
extension and its underlying mathematical
framework (Subsection~\ref{sub_sec:rmtgb}).
Finally, a theoretical analysis is presented
(Subsection~\ref{subsec:theoretical_analysis}).

\subsection{Preliminaries and Notation}
\label{sub_sec:preliminaries}

In this study, we define the input space
as \(\mathcal{X} \subseteq \mathbb{R}^{d}\),
where each input \(\textbf{x} \in \mathcal{X}\) is a
\(d\)-dimensional
feature vector. The corresponding output space
is \(\mathcal{Y} \subseteq \mathbb{R}\),
where each output \(y \in \mathcal{Y}\) is a
target scalar value, in the case of regression problems.
In the case of classification we consider a one-hot encoding
scheme for the targets, \emph{i.e.}, \(\mathcal{Y} \subset \{0,1\}^K\),
with $K$ the number of classes.
The dataset is denoted by
\(\mathcal{D} = \{(\mathbf{x}_i, y_i)\}_{i=1}^N\),
where each sample \((\mathbf{x}_i, y_i)\)
is drawn independently and identically
distributed (i.i.d.) from \(P(\mathcal{X}, \mathcal{Y})\),
and \(N\) is the number of samples.
This corresponds to a supervised learning setting,
where the goal is to learn a mapping from inputs to outputs
using labeled data~\citep{Cunningham2008}.

Subsequently, to evaluate the performance of the training model,
we define a loss function
\(\mathcal{L}(y, \hat{F})\), which measures the discrepancy
between the true output
\(y\) and the model's output \(\hat{F}\).
The specific form of the loss function
depends on the nature of the problem~\citep{Wang2022loss}.
In this study, we use the cross-entropy loss
function for classification,
\begin{equation}
      \label{eq:cross_entropy}
      \mathcal{L}(\mathbf{y}, \mathbf{\hat{F}}) =
      -\sum_{k=1}^{K} y_k \ln(P_k)\,,
\end{equation}
where \(K\) is the number of distinct class labels,
\(\mathbf{y}\) is a one-hot encoded vector of length $K$, and
\(P_k\) is the predicted probability of class \(k\),
\begin{equation}
      \label{eq:pred_proba}
      P_k = \frac{\exp (\hat{F}_k)}{\sum_{k=1}^{K} \exp (\hat{F}_k)}\,.
\end{equation}
Furthermore, $\mathbf{\hat{F}}$ is a vector of length $K$ with
the specific model's output for each of the $K$ class labels
associated to the corresponding inputs $\mathbf{x}$.

For regression, we employ the squared error loss function,
\begin{equation}
      \label{eq:squared_error}
      \mathcal{L}(y, {\hat{F}}) =\frac{1}{2}
      \left(y - {\hat{F}}\right)^2\,,
\end{equation}
where $y$ is the observed target and $\hat{F}$
is the model's output for the input $\mathbf{x}$.

\subsection{Multi-Task Learning}
\label{sub_sec:mt}
Considering a collection of \( T \) tasks,
each task \( t \in \{1, \ldots, T\} \) is associated with its
own input-output space,
\( \mathcal{X}_{(t)} \) and
\( \mathcal{Y}_{(t)} \), respectively.
We assume a shared input space
\(\mathcal{X} = \mathcal{X}_{1} = \cdots = \mathcal{X}_{T}\),
with consistent input feature dimensionality \(d^{(t)} = d\)
across all tasks.
Similarly, the output space is shared among tasks,
\(\mathcal{Y} = \mathcal{Y}_{1} = \cdots = \mathcal{Y}_{T}\).
Each task \( t \) has its own dataset,
\begin{equation}
      \label{eq:mtl_dataset}
      \mathcal{D}_{t} = \{(\mathbf{x}_{i,t}, y_{i,t})\}_{i=1}^{N^{(t)}},
\end{equation}
where \( (\mathbf{x}_{i,t}, y_{i,t})
\sim P_{t}(\mathcal{X}_{t}, \mathcal{Y}_{t}) \).
While these tasks are generally assumed to be
related~\citep{Caruana1997,Olivier2011,Emami2023},
in practice, the collection may contain \emph{outlier tasks}
that deviate significantly from the shared (common)
structure.
Such tasks can negatively impact the quality of the
shared representation and degrade overall performance
if treated uniformly within the learning
process~\citep{Gong2012,Yu2007}.

The goal of \ac{MTL} is to simultaneously learn a collection of
task-specific functions~\citep{Evgeniou2004},
\begin{align*}
      \{F_t : \mathcal{X}_{t} \rightarrow \mathcal{Y}_{t}\}_{t=1}^T,
\end{align*}
that collectively minimize the total loss across all tasks,
\begin{equation}
      \label{eq:mtl_objective}
      F(\mathbf{x})= \argmin_{\{ \hat{F}_{t} \}_{t=1}^{T}}
      \sum_{t=1}^{T} \sum_{i=1}^{N^{(t)}}
      \bigg[\mathcal{L}\left(y_{i,t}, \hat{{F}}_{t}
            (\mathbf{x}_{i,t})\right)\bigg].
\end{equation}

To facilitate parameter sharing across tasks,
\ac{MTL} models can alternatively express each task-specific
function \( F_{t} \)
as the sum of a shared component and a task-specific component,
\begin{equation}
      \label{eq:shared_additive}
      F_t(\mathbf{x}) = \phi(\mathbf{x}) + \psi_t(\mathbf{x}),
\end{equation}
where \( \phi: \mathcal{X} \rightarrow \mathcal{Y} \) denotes a
\emph{shared function}
capturing common structure across tasks,
and \( \psi_{t}: \mathcal{X}_{t} \rightarrow \mathcal{Y}_{t} \)
is a \emph{task-specific function}
modeling individual task characteristics.
This additive formulation enables the model to learn a
global inductive bias via \( \phi \),
while still allowing per-task flexibility through \( \psi_{t} \).
However, the presence of outlier tasks,
which do not align well with the dominant task structure,
can mislead the learning of the shared representation \( \phi \),
resulting in degraded performance across the entire task set.

\subsection{Gradient Boosting}
\label{sub_sec:gb}
The primary objective of \ac{GB} model,
as introduced by~\cite{Friedman2001}, is to iteratively
minimize a given loss
\( \mathcal{L}\left(y, \hat{F}(\mathbf{x})\right) \),
by finding a function that maps the input features
\(\mathbf{x}\) to the predicted output \(\hat{F}\),

\begin{equation}
      \label{eq:gb_objective}
      F(\mathbf{x}) = \argmin_{\hat{F}(\mathbf{x})}
      \sum_{i=1}^{N}
      \bigg[\mathcal{L}\left(y_i, \hat{F}(\mathbf{x}_i)\right)\bigg].
\end{equation}

This optimization is performed forward stage-wise
by sequentially adding base learners
\(h_m(\mathbf{x}) =\)
and incorporating
the ensemble parameter \(\gamma\)
at each boosting iteration \(m\) to the model,

\begin{equation}
      \label{eq:gb_additive_model}
      \hat{F}_M(\mathbf{x}) = \sum_{m=0}^{M} \gamma_m h_m(\mathbf{x}).
\end{equation}
The model is initialized with a constant
value that minimizes the loss,
\begin{equation}
      \label{eq:gb_F0}
      \hat{F}_0(\mathbf{x}) = \argmin_\gamma \sum_{i=1}^{N}
      \mathcal{L}(y_i, \gamma).
\end{equation}
Hence, Eq.~\eqref{eq:gb_objective} can be expressed as
a stage-wise greedy process,
\begin{equation}
      \label{eq:gb_objective_parameterized}
      (\gamma_m, h_m) = \argmin_{\{\gamma_m, h_m\}}
      \sum_{i=1}^{N}
      \mathcal{L}\left(y_i, \hat{F}_{m-1}(\mathbf{x}_i) +
      \gamma_m h_m(\mathbf{x}_i)
      \right).
\end{equation}

At each iteration $m$, instead of directly
optimizing Eq.~\eqref{eq:gb_objective_parameterized},
\ac{GB} utilizes the negative gradient of the loss function
(pseudo-residuals) with respect
to the prediction of the current model to guide the learning
of the next base learner,
\begin{equation}
      \label{eq:pseudo}
      r_{i, m} = -\left[\frac{\partial
                  \mathcal{L}(y_i, \hat{F}(\mathbf{x}_i))}
            {\partial \hat{F}(\mathbf{x}_i)}\right]
      _{F=\hat{F}_{m-1}(\mathbf{x}_i)},
\end{equation}
for each sample \(i\) in the dataset.
A new base learner \( h_m(\mathbf{x}) \) is then fitted
to these pseudo-residuals by minimizing the squared error
(regardless of the loss function the ensemble is trying to optimize),
\begin{equation}
      \label{eq:tree_stump_fit}
      h_m(\mathbf{x}) = \argmin_{\{h \in \mathcal{H}\}}
      \sum_{i=1}^{N} \left(r_{i,m} - h_m(\mathbf{x}_i)\right)^2,
\end{equation}
where \(\mathcal{H}\) denotes the hypothesis space of base
learners, typically a set of \acf{DTR}.
Once the base learner \( h_m(\mathbf{x}) \) is determined,
the optimal parameter \( \gamma_m \) is obtained by solving the
line search problem,
\begin{equation}
      \label{eq:line_search}
      \gamma_m = \argmin_{\gamma_m}
      \sum_{i=1}^{N} \mathcal{L}\left(y_i, \hat{F}_{m-1}(\mathbf{x}_i) +
      \gamma_m h_m(\mathbf{x}_i)\right).
\end{equation}
However, in practice, often $\gamma_m$ is set simply equal to $1$.

After \( M \) boosting iterations, the final predictive model is
built as an additive ensemble of base learners,
\begin{equation}
      \label{eq:final_model}
      F(\mathbf{x}) = \hat{F}_M(\mathbf{x}) =
      \hat{F}_0(\mathbf{x}) +
      \eta \sum_{m=1}^{M} \gamma_m h_m(\mathbf{x}),
\end{equation}
where \(\eta \in (0, 1]\) is a learning rate, that is used to
regularize the gradient descent steps in the learning process.
In summary, \ac{GB} simply performs gradient descent in function
space by incorporating, at each boosting iteration, a new predictor into the ensemble.

\ac{GB} has consistently shown strong empirical performance
across a diverse set of \ac{ML} problems,
including regression~\citep{Samir2018,Wenchao2024},
binary and multi-class
classification~\citep{Taha2020,Gunasekara2024},
ranking~\citep{Plaia2022},
missing value estimation~\citep{Manar2022},
and \ac{MTL}
problems~\citep{Olivier2011,Liu_Haizhou2024,Emami2023}.
Its flexibility in accommodating different loss
functions makes it well-suited for both standard and
specialized applications~\citep{Natekin2013,Bentejac2021}.

\subsection{Robust Multi-Task Gradient Boosting}
\label{sub_sec:rmtgb}
To address challenges in \ac{MTL},
such as task heterogeneity and outlier task influence,
we propose a three-stage \ac{GB}
framework called \ac{R-MTGB},
which integrates robustness and shared representation
learning within the \ac{GB} paradigm.
The training process of \ac{R-MTGB}
model is divided into three sequential blocks,
where each block has a specific motivation
that progressively refines the learning process:
(i) initialize with general knowledge,
(ii) enforce robustness against outliers, and
(iii) specialize to individual tasks.

\begin{itemize}
      \item \textbf{Block 1 (Shared Representation Learning).}
            Focuses on shared representation
            learning by leveraging all tasks jointly to identify a
            common latent function that captures task-invariant patterns.
            This prevents \emph{cold-start} bias by providing
            a strong initialization before any task-specific
            adaptation.

      \item \textbf{Block 2 (Outlier-Aware Partitioning).}
            Introduces robustness to task outliers by distinguishing between
            non-outlier and outlier tasks through a sigmoid-based weighting
            mechanism. This mechanism assigns extreme weights to outlier tasks,
            amplifies the contribution of reliable tasks, and suppresses the
            influence of misaligned ones, thereby enabling the model to focus
            on the most informative task signals and mitigate negative transfer.

      \item \textbf{Block 3 (Task-Specific Fine-Tuning).}
            Performs task-specific refinement,
            where individual models are fine-tuned for each
            task based on the previously learned shared
            and robust representations.
            This allows the recovery of fine-grained
            task details that joint training may suppress.
\end{itemize}

Each block builds upon the outputs of the previous blocks,
progressively refining the model to improve performance
across both related and unrelated tasks. Formally,
the total number of boosting iterations
$M$ is partitioned into three phases:
\begin{align*}
      M = M_1 + M_2 + M_3,
\end{align*}
where \( M_1 \), \( M_2 \), and \( M_3 \) correspond to
the boosting iterations assigned to Block 1, Block 2, and Block 3,
respectively. The number of iterations of each block is a hyperparameter
that will be adjusted in practice using a cross-validation grid search.

The final proposed ensemble prediction function
for a given input \( \mathbf{x} \) is
\begin{equation}
      \label{eq:rmtgb_ensemble}
      \begin{split}
            F_{t}(\mathbf{x}) =
            \hat{F}^{(\mathrm{shared})}(\mathbf{x}) +
            \bigl(1 - \sigma(\theta_{t})\bigr)\cdot
            \hat{F}^{(\mathrm{non\mbox{-}outlier})}(\mathbf{x}) \\
            + \sigma(\theta_{t}) \cdot \hat{F}^{(\mathrm{outlier})}
            (\mathbf{x}) + \hat{F}_t^{(\mathrm{task})}(\mathbf{x})
            ,
      \end{split}
\end{equation}
where,
\sloppy
\begin{itemize}
      \item \(\hat{F}^{(\mathrm{shared})}\) is the shared-model
            that captures global shared structures across all tasks.
      \item \(\hat{F}^{(\mathrm{non\mbox{-}outlier})}\) models
            patterns characteristic of non-outlier tasks.
      \item \(\hat{F}^{(\mathrm{outlier})}\) captures patterns
            specific to outlier tasks.
      \item \(\hat{F}_{t}^{(\mathrm{task})}\) represents the
            task-specific fine-tuned model
            for task $t$.
      \item \(\sigma(\theta_{t}) = \frac{1}{1 + \exp({-\theta_{t}})}\)
            is the sigmoid function that simply outputs the probability that
            a tasks is an outlier  tasks.
\end{itemize}
Although the ensemble prediction function
in Eq.~\eqref{eq:rmtgb_ensemble} contains four components,
they are learned iteratively through 3 training blocks. Specifically,
Block 2 \emph{jointly} models both the outlier and non-outlier tasks via
a unified regularization mechanism that allocates task-specific weights.
This shared optimization process gives rise to two separate
components in the second block. Namely, \(\hat{F}^{(\mathrm{outlier})}\) and \(\hat{F}^{(\mathrm{non\mbox{-}outlier})}\).
Importantly, each individual function is obtained by combining the different base learners
generated at each block. Namely,
\begin{equation}
      \hat{F}^{(\mathrm{shared})}(\mathbf{x})
      = \sum_{m=1}^{M_1} \eta\, h_m^{(\mathrm{shared})}(\mathbf{x})\,.
      \label{eq:f_shared}
\end{equation}

\begin{equation}
      \hat{F}^{(\mathrm{outlier})}(\mathbf{x})
      = \sum_{m=1}^{M_2} \eta\, h_m^{(\mathrm{outlier})}(\mathbf{x})\,.
      \label{eq:f_outlier}
\end{equation}

\begin{equation}
      \hat{F}^{(\mathrm{non\text{-}outlier})}(\mathbf{x})
      = \sum_{m=1}^{M_2} \eta\, h_m^{(\mathrm{non\text{-}outlier})}(\mathbf{x})\,.
      \label{eq:f_nonoutlier}
\end{equation}

\begin{equation}
      \hat{F}_t^{(\mathrm{task})}(\mathbf{x})
      = \sum_{m=1}^{M_3} \eta\, h_{m,t}^{(\mathrm{task})}(\mathbf{x})\,.
      \label{eq:f_task}
\end{equation}
where $\eta$ is the learning rate considered and each base learner,
denoted by \( h_m^{(\cdot)} \) and \( h_{m,t}^{(\cdot)} \), is implemented
as a \ac{DTR}. In the case of multi-class classification
\(h_m^{(\cdot)}\) and \(h_{m,t}^{(\cdot)}\)
are multi-output \ac{DTR}s, as in~\cite{Emami2025}.
Class probabilities are simply obtained by applying the soft-max
activation function.

These base learners are trained in sequence, in an iterative process, at
each block, by fitting the pseudo-residuals
associated to the current state of the corresponding individual function
\(\hat{F}^{(\mathrm{shared})}\), \(\hat{F}^{(\mathrm{non\mbox{-}outlier})}\)
\(\hat{F}^{(\mathrm{outlier})}\) or \(\hat{F}_{t}^{(\mathrm{task})}\).
This process is equivalent to performing gradient
descent in function space, as in the standard \ac{GB} algorithm.
More precisely, the corresponding objective that
is minimized to fit each \( h_m^{(\cdot)} \) and each \( h_{m,t}^{(\cdot)} \), at each block, is
\begin{align}
      \mathcal{L}_m^{(\mathrm{shared})}      & =
      \sum_{t=1}^{T} \sum_{i=1}^{N_t} ||h_m^{(\mathrm{shared})}(\mathbf{x}_{i,t}) -
      r_{i,m,t}^{(\mathrm{shared})}||_2^2\,, \label{eq:loss_hm_shared}            \\
      \mathcal{L}_m^{(\mathrm{outlier})}     & =
      \sum_{t=1}^{T} \sum_{i=1}^{N_t} ||h_m^{(\mathrm{outlier})}(\mathbf{x}_{i,t}) -
      r_{i,m,t}^{(\mathrm{outlier})}||_2^2\,, \label{eq:loss_hm_outlier}          \\
      \mathcal{L}_m^{(\mathrm{non-outlier})} & =
      \sum_{t=1}^{T} \sum_{i=1}^{N_t} ||h_m^{(\mathrm{non-outlier})}(\mathbf{x}_{i,t}) -
      r_{i,m,t}^{(\mathrm{non-outlier})}||_2^2 \,, \label{eq:loss_hm_non_outlier} \\
      \mathcal{L}_{m,t}^{(\mathrm{task})}    & =
      \sum_{i=1}^{N_t} ||h_{m,t}^{(\mathrm{task})}(\mathbf{x}_{i,t}) -
      r_{i,m,t}^{(\mathrm{task})}||_2^2 \,, \label{eq:loss_hm_task}
\end{align}
where \( \mathbf{x}_{i,t} \) is the input for instance \( i \)
in task \(t\) and \( r_{i,m, t}^{(\cdot)} \) is the
corresponding pseudo-residual at iteration \( m \).

The pseudo-residuals \( r_{i,m,t}^{(\cdot)} \)
are recalculated at every boosting iteration \( m \) for their
respective block, for the corresponding number of iterations \( M_1, M_2, \)
and \( M_3 \).
During Block~2, the pseudo-residuals are weighted by a task-specific factor
$\theta_t$ reflecting task relatedness, so that outlier tasks have less
influence on the shared component. The weights $\theta_t$ are learned
jointly during Block~2, as described below.
In our implementation, the initial ensemble prediction in
Eq.~\eqref{eq:rmtgb_ensemble} is set to zero before any learning occurs.
The following paragraphs describe each block in detail.

\textbf{Block 1 - shared-Learning via Data Pooling}: In the first block,
\(M_1\) shared-level predictors are trained, iteratively, to form a
shared set of base learners
\( h_{m}^{\mathrm{(shared)}}(\mathbf{x}_i, r_{i,m,t}^{(\mathrm{shared})})\)
that constitute $\hat{F}^{(\mathrm{shared})}(\mathbf{x})$,
as indicated in Eq. (\ref{eq:f_shared}),
using pooled data from all tasks,
\begin{align}
      \label{eq:rmtgb_pool}
      \mathcal{D}_{\mathrm{pool}} = \bigcup_{t=1}^{T} \mathcal{D}^{(t)},
\end{align}
and pseudo-residuals computed as,
\begin{align}
      \label{eq:rmtgb_re_shared}
      r_{i,m,t}^{(\mathrm{shared})}
       & = - \frac{\partial \mathcal{L}(y_{i,t}, F(\mathbf{x}_{i,t}))}
      {\partial \hat{F}^{(\mathrm{shared})}(\mathbf{x}_{i,t})} \notag  \\
       & = - \frac{\partial \mathcal{L}(y_{i,t}, F(\mathbf{x}_{i,t}))}
      {\partial F(\mathbf{x}_{i,t})}
      \cdot \frac{\partial F(\mathbf{x}_{i,t})}
      {\partial \hat{F}^{(\mathrm{shared})}(\mathbf{x}_{i,t})} \notag  \\
       & = - \frac{\partial \mathcal{L}(y_{i,t}, F(\mathbf{x}_{i,t}))}
      {\partial F(\mathbf{x}_{i,t})}
\end{align}
That is, the pseudo-residuals in this block are vectors pointing
in the negative gradient of the loss
with respect to $\hat{F}^{(\mathrm{shared})}$.
Once a predictor \( h_{m}^{\mathrm{(shared)}} \)
has been fitted by minimizing Eq. (\ref{eq:loss_hm_shared}),
it is incorporated into $\hat{F}^{(\mathrm{shared})}$.
This process repeats for $M_1$ iterations.

\textbf{Block 2 - Outlier-Aware Task Partitioning}:
To mitigate the impact of task outliers,
the second block adopts a two-branch structure
over the pooled data (\(\mathcal{D}_{\mathrm{pool}}\)).
One branch targets outlier tasks, while the other focuses on
non-outlier tasks.  The outlier tasks branch is obtained by fitting
\( h_{m}^{(\mathrm{outlier})}
(\mathbf{x}_i,r_{i,m,t}^{(\mathrm{outlier})})\),
to the
negative gradient of the loss with respect to \(  F^{(\mathrm{outlier})}
\),
\begin{align}
      \begin{aligned}
            \label{eq:rmtgb_chain_outlier}
            r_{i,m,t}^{(\mathrm{outlier})}
             & =
            - \frac{\partial \mathcal{L}(y_{i,t}, F(\mathbf{x}_{i,t}))}
            {\partial \hat{F}^{(\mathrm{outlier})}(\mathbf{x}_{i,t})} \\
             & =
            - \frac{\partial \mathcal{L}(y_{i,t}, F(\mathbf{x}_{i,t}))}
            {\partial F(\mathbf{x}_{i,t})}
            \cdot \frac{\partial F(\mathbf{x}_{i,t})}
            {\partial \hat{F}^{(\mathrm{outlier})}(\mathbf{x}_{i,t})},
      \end{aligned}
\end{align}
which yields,
\begin{equation}
      \label{eq:rmtgb_re_2_final}
      r_{i,m,t}^{(\mathrm{outlier})} = -
      \frac{\partial \mathcal{L}(y_{i,t}, F(\mathbf{x}_{i,t}))}
      {\partial F(\mathbf{x}_{i,t})}
      \cdot \sigma({\theta}_t).
\end{equation}
That is, the pseudo-residuals in this branch of the second block
(Block 2a) are vectors
pointing in the negative gradient of the loss
with respect to $\hat{F}^{(\mathrm{outlier})}$.
Once a predictor \( h_{m}^{\mathrm{(outlier)}} \)
has been fitted by minimizing Eq. (\ref{eq:loss_hm_outlier}), it is
incorporated into $\hat{F}^{(\mathrm{outlier})}$.

In a similar manner, the non-outlier branch (Block 2b) is obtained by fitting
\( h_{m}^{(\mathrm{non\text{-}outlier})} \)
to the negative gradient of the loss with respect to
\( F^{(\mathrm{non\text{-}outlier})} \)
\begin{align}
      \begin{aligned}
            \label{eq:rmtgb_chain_non_outlier}
            r_{i,m,t}^{(\mathrm{non\text{-}outlier})}
             & =
            - \frac{\partial \mathcal{L}(y_{i,t}, F(\mathbf{x}_{i,t}))}
            {\partial \hat{F}^{(\mathrm{non\text{-}outlier})}(\mathbf{x}_{i,t})} \\
             & =
            - \frac{\partial \mathcal{L}(y_{i,t}, F(\mathbf{x}_{i,t}))}
            {\partial F(\mathbf{x}_{i,t})}
            \cdot \frac{\partial F(\mathbf{x}_{i,t})}
            {\partial \hat{F}^{(\mathrm{non\text{-}outlier})}(\mathbf{x}_{i,t})},
      \end{aligned}
\end{align}
which gives,
\begin{equation}
      \label{eq:rmtgb_re_3_final}
      r_{i,m,t}^{(\mathrm{non\text{-}outlier})}
      = -
      \frac{\partial \mathcal{L}(y_{i,t}, F(\mathbf{x}_{i,t}))}
      {\partial F(\mathbf{x}_{i,t})}
      \cdot (1 - \sigma({\theta}_t)).
\end{equation}
That is, the pseudo-residuals in this branch of the second block
(Block 2b) are vectors
pointing in the negative gradient of the loss
with respect to $\hat{F}^{(\mathrm{non\text{-}outlier})}$.
Once a base learner \( h_{m}^{\mathrm{(non\text{-}outlier)}} \)
has been fitted by minimizing Eq. (\ref{eq:loss_hm_non_outlier}),
it is incorporated into
$\hat{F}^{(\mathrm{non\text{-}outlier})}$.
These two steps, described to update $\hat{F}^{(\mathrm{non\text{-}outlier})}$
and $\hat{F}^{(\mathrm{outlier})}$, are repeated for $M_2$ iterations.

After generating each \( h_{m}^{\mathrm{(outlier)}} \) and each
\( h_{m}^{\mathrm{(non\text{-}outlier)}} \)
in Block 2, the parameter $\theta_t$, for each task is also updated.
This dynamically adjusts the influence of each task $t$ in the final
ensemble prediction through a
sigmoid-based weighting mechanism (See Eq. (\ref{eq:rmtgb_ensemble})).

Rather than explicitly labeling tasks as outliers or non-outliers,
the model is \emph{encouraged to learn} a soft partitioning,
where the sigmoid activations of $\theta_t$ tend towards—but do
not always reach—extreme values close to 0 or 1.
This soft activation enables flexible modulation of the
contribution of each task to the outlier and non-outlier components.
By doing so, the model can reduce the influence of
anomalous outlier tasks, that may impair the \ac{MTL} process,
while emphasizing signals
from more consistent tasks. As training progresses,
negative gradients of the loss function \(\mathcal{L}\) with respect
to parameter vector $\boldsymbol{\theta}$ guide this modulation,
allowing the optimization process to adaptively infer and
separate outlier tasks in a data-driven manner.

The update of $\boldsymbol{\theta}$ is done by taking a step of
size $\eta$ (\emph{i.e.}, the learning rate) in the negative direction
of the loss gradient with respect to $\boldsymbol{\theta}$. That is,
\begin{align}
      \begin{aligned}
            \label{eq:grad_theta}
            -\frac{\partial \mathcal{L}}{\partial \boldsymbol{\theta}}
             & =
            -\frac{\partial \mathcal{L}}{\partial F} \cdot
            \frac{\partial F}{\partial \boldsymbol{\theta}} \\
             & =
            -\sum_{t=1}^{T} \sum_{i=1}^{N^{(t)}}
            \frac{\partial \mathcal{L}\left(y_{i,t},
                  F(\mathbf{x}_{i,t})\right)}
            {\partial F(\mathbf{x}_{i,t})}
            \cdot
            \frac{\partial F(\mathbf{x}_{i,t})}{\partial
                  {\theta}_t},
      \end{aligned}
\end{align}
where,
\begin{align}
      \label{eq:partial_obj_theta}
      -\frac{\partial F(\mathbf{x}_{i,t})}{\partial {\theta_t}}
       & =
      -\sigma({\theta_t}) \cdot
      \bigl(1 - \sigma({\theta_t})\bigr) \cdot
      \, \hat{F}^{(\mathrm{non\text{-}outlier})}(\mathbf{x}_{i,t}) \notag \\
       & \quad +
      \sigma({\theta_t}) \cdot \bigl(1 - \sigma({\theta_t})\bigr)
      \, \cdot \hat{F}^{(\mathrm{outlier})}(\mathbf{x}_{i,t}).
\end{align}
Consequently, the negative gradient of the loss with
respect to the parameter \( \theta_{t} \)
(Eq.~\eqref{eq:grad_theta}) is computed as,
\begin{equation}
      \label{eq:gradient_theta}
      \begin{aligned}
            -\frac{\partial \mathcal{L}}{\partial \theta_t}
            = \, & \sum_{t=1}^{T} \sum_{i=1}^{N^{(t)}} r_{i,m,t}^{(\mathrm{shared})} \cdot \sigma\bigl(\theta_t\bigr)
            \cdot \bigl(1 - \sigma\bigl(\theta_t\bigr)\bigr)                                                          \\
                 & \quad \cdot
            \left[ \hat{F}_{t}^{(\mathrm{outlier})}\bigl(\mathbf{x}_{i,t}\bigr)
            - \hat{F}_{t}^{(\mathrm{non\text{-}outlier})}\bigl(\mathbf{x}_{i,t}\bigr) \right]\,.
      \end{aligned}
\end{equation}

\textbf{Block 3 - Task-Specific Fine-Tuning}:
This block operates as a standard \acf{ST}-\ac{GB},
initialized with the learned functions from previous blocks,
$\hat{F}^{(\mathrm{shared})}$, $\hat{F}^{(\mathrm{outlier})}$ and $\hat{F}^{(\mathrm{non\text{-}outlier})}$.
Specifically, we simply update each  \(\hat{F}_{t}^{(\mathrm{task})}\) in Eq. (\ref{eq:rmtgb_ensemble}).
For this, each task independently fits
\( h_m^{(\mathrm{task})}(\mathbf{x}_i, r_{i,m,t}^{(\mathrm{task})}) \),
to the corresponding pseudo-residuals,
\begin{align}
      \label{eq:rmtgb_re_task}
      r_{i,m,t}^{(\mathrm{task})}
       & =
      - \frac{\partial \mathcal{L}(y_{i,t}, F(\mathbf{x}_{i,t}))}
      {\partial \hat{F}^{(\mathrm{task})}(\mathbf{x}_{i,t})} \notag \\
       & =
      - \frac{\partial \mathcal{L}(y_{i,t}, F(\mathbf{x}_{i,t}))}
      {\partial F(\mathbf{x}_{i,t})}
      \cdot \frac{\partial F(\mathbf{x}_{i,t})}
      {\partial \hat{F}^{(\mathrm{task})}(\mathbf{x}_{i,t})} \notag \\
       & =
      -
      \frac{\partial \mathcal{L}(y_{i,t}, F(\mathbf{x}_{i,t}))}
      {\partial F(\mathbf{x}_{i,t})}.
\end{align}
That is, the pseudo-residuals computed in Block 3 are
vectors pointing in the negative gradient of the loss
with respect to $\hat{F}_t^{(\mathrm{task})}$, for each task $t$.
Once a base learner \( h_{m,t}^{\mathrm{(task)}} \)
has been fitted by minimizing Eq. (\ref{eq:loss_hm_task}),
it is incorporated into $\hat{F}_t^{(\mathrm{task})}$.
This step is repeated for $M_3$ iterations, for each task.
The goal of this block is hence to allow each task to capture unique patterns not
shared by previous blocks.

Note that the computation of the pseudo-residuals involves the evaluation of ${-\partial \mathcal{L}(y_{i,t}, F(\mathbf{x}_{i,t}))} /
      {\partial F(\mathbf{x}_{i,t})}$.
Depending on the choice of the loss
function for each problem type
(\emph{i.e.}, classification or regression),
these gradients are computed differently.
For classification, using the cross-entropy loss
in Eq.~\eqref{eq:cross_entropy}, the gradients are defined as
the difference between the true labels and predicted
probabilities specified in Eq.~\eqref{eq:pred_proba}. That is,
\begin{equation}
      \label{eq:residual_classification_summation}
      \frac{-\partial \mathcal{L}(y_{i,t}, F(\mathbf{x}_{i,t}))}
      {\partial F(\mathbf{x}_{i,t})}= y_{i,k} - P_{i,k}(\mathbf{x}_{i}).
\end{equation}
In the case of regression, using the
squared error loss defined in Eq.~\eqref{eq:squared_error},
the gradients are given by the difference
between the true and predicted outputs,
\begin{equation}
      \label{eq:residual_regression}
      \frac{-\partial \mathcal{L}(y_{i,t}, F(\mathbf{x}_{i,t}))}{\partial F(\mathbf{x}_{i,t})} = y_i - \hat{F}(\mathbf{x}_{i}).
\end{equation}

The training procedure of \ac{R-MTGB} is summarized in
algorithm~\ref{alg:rmtgb}.

\begin{algorithm}[H]
      \caption{\acf{R-MTGB} Training Procedure.}
      \label{alg:rmtgb}
      \KwIn{$\{\mathcal{D}_{t}\}_{t=1}^T, \mathcal{L}, M=M_1+M_2+M_3,
            \eta \in (0, 1], \theta \sim \mathcal{N}(\mu,\sigma^2)$}
      \KwOut{$F(\mathbf{x})$}
      \BlankLine
      Initialize: \quad
      $\hat{F}^{\mathrm{shared}}_0 =
            \hat{F}^{\mathrm{outlier}}_0 =
            \hat{F}^{\mathrm{non\mbox{-}outlier}}_0 =
            \hat{F}^{\mathrm{task}}_0 = 0,
            \quad \mathcal{D}_{\mathrm{pool}} = \bigcup_{t=1}^T
            \mathcal{D}^{(t)}$
      \BlankLine
      \textbf{Block 1: shared-Learning}\;
      \For{$m=1$ \KwTo $M_1$}{
      $\forall t,i: r_{i,m,t}^{(shared)} \leftarrow $ Eq.~\eqref{eq:rmtgb_re_shared}\;
      $h_m^{(shared)} \leftarrow \texttt{fit}(\mathcal{D}_{pool}, r_{m}^{(shared)})$\;
      $\hat{F}^{(shared)}_m \leftarrow \hat{F}^{(shared)}_{m-1} + \eta h_m^{(shared)}$
      }
      \BlankLine
      \textbf{Block 2: Outlier Partitioning}\;
      \For{$m=(M_1 + 1)$ \KwTo $(M_1 + M_2)$}{
      $\forall t, i: \quad
            r_{i,m,t}^{(\text{outlier})} \leftarrow \text{Eq.}~\eqref{eq:rmtgb_re_2_final}, \quad
            r_{i,m,t}^{(\mathrm{non\mbox{-}outlier})} \leftarrow
            \text{Eq.}~\eqref{eq:rmtgb_re_3_final}$\;

      $h_m^{(outlier)} \leftarrow \texttt{fit}(\mathcal{D}_{pool}, r_{m}^{(outlier)}), \quad
            \hat{F}^{(outlier)}_m \leftarrow \hat{F}^{(outlier)}_{m-1} + \eta h_m^{(outlier)}$\;
      $h_m^{(\mathrm{non\mbox{-}outlier})} \leftarrow
            \texttt{fit}(\mathcal{D}_{pool}, r_{m}^{(\mathrm{non\mbox{-}outlier})}), \quad
            \hat{F}^{(\mathrm{non\mbox{-}outlier})}_m \leftarrow
            \hat{F}^{(\mathrm{non\mbox{-}outlier})}_{m-1} + \eta h_m^{\mathrm{non\mbox{-}outlier}}$;

      $\forall t: {\theta^*_{m,t}} \leftarrow \theta_{(m-1), t} -
            \eta \frac{\partial \mathcal{L}}
            {\partial \theta_{(m-1), t}}$ using Eq.~\eqref{eq:gradient_theta}\;

      }
      \BlankLine
      \textbf{Block 3: Task-Specific Fine-Tuning}\;
      \For{$m = (M_1 + M_2 + 1)$ \KwTo $M$}{
            \For{$t=1$ \KwTo $T$}{
                  $\forall i: r_{i,m, t}^{(task)} \leftarrow $\ Eq.~\eqref{eq:rmtgb_re_task}\;
                  $h_{m,t}^{(task)} \leftarrow \texttt{fit}(\mathcal{D}_{t}, r_{m,t}^{(task)})$\;
                  $\hat{F}_{m, t}^{(task)} \leftarrow \hat{F}_{(m-1), t}^{(task)} + \eta h_{m}^{(task),(t)}$
            }
      }
      \BlankLine
      \KwRet{$\hat{F}^{(shared)}(\mathbf{x}) +
            (1-\sigma(\boldsymbol{\theta})) \cdot \hat{F}^{(\mathrm{non\mbox{-}outlier})}(\mathbf{x}) +
            \sigma(\boldsymbol{\theta})\cdot \hat{F}^{(outlier)}(\mathbf{x}) +
            \hat{F}^{(task)}(\mathbf{x})$}
\end{algorithm}

\subsection{Theoretical Analysis of Block 2}
\label{subsec:theoretical_analysis}

Block~2 guarantees that the contribution of each task to the empirical loss
is bounded by a sigmoid weight, ensuring that outlier tasks cannot dominate
the optimization. From Eq.~\eqref{eq:rmtgb_re_2_final}, and
Eq.~\eqref{eq:rmtgb_re_3_final}, the pseudo-residuals are multiplied by
$\sigma(\theta)$ and $(1-\sigma(\theta))$, so for
every task $t$ and sample $i$,
\begin{equation}
      \big| r_{i,m,t}^{(\mathrm{outlier})} \big|
      \;\leq\; \big| r_{i,m,t}^{(\mathrm{shared})} \big|,
      \qquad
      \big| r_{i,m,t}^{(\mathrm{non\text{-}outlier})} \big|
      \;\leq\; \big| r_{i,m,t}^{(\mathrm{shared})} \big|.
\end{equation}
To optimize the task-specific weights $\theta_t$, the model computes the
negative gradient of the empirical risk with respect to $\theta_t$ (see
Eq.~\eqref{eq:gradient_theta})
\begin{equation}
      \label{eq:gradient_theta_with_bound}
      - \frac{\partial \mathcal{L}}{\partial \theta_t}
      = \sigma(\theta_t)\bigl(1-\sigma(\theta_t)\bigr)\,S_t,
\end{equation}
where,
\begin{equation}
      S_t = \sum_{i=1}^{N_t} r_{i,m,t}^{(\mathrm{shared})}
      \Bigl(\hat{F}_t^{(\mathrm{outlier})}\bigl(\mathbf{x}_{i,t}
      \bigr) - \hat{F}_t^{(\mathrm{non\text{-}outlier})}
      \bigl(\mathbf{x}_{i,t}\bigr)\Bigl).
\end{equation}
Because $\sigma(z)\in[0,1]$ for all $z$, the product
$\sigma(z)(1-\sigma(z))$ is always non-negative and is maximized at
$z=0$, with
\[
      0 \;\leq\; \sigma(z)(1-\sigma(z)) \;\leq\; \tfrac{1}{4},
      \quad \forall z \in \mathbb{R}.
\]
Therefore, the Eq.~\eqref{eq:gradient_theta_with_bound}
is bounded as,
\begin{equation}
      \Bigg| -\frac{\partial \mathcal{L}}{\partial \theta_t} \Bigg|
      \;\leq\; \frac{1}{4} \,
      \sum_{i=1}^{N_t} \big| r_{i,m,t}^{(\mathrm{shared})} \big| \,
      \Big| \hat{F}_t^{(\mathrm{outlier})}\bigl(\mathbf{x}_{i,t}
      \bigr) - \hat{F}_t^{(\mathrm{non\text{-}outlier})}
      \bigl(\mathbf{x}_{i,t}\bigr) \Big|.
\end{equation}
This upper bound ensures stable updates and prevents extreme tasks
from overwhelming the optimization.

\paragraph{Proposition 1 (Task-wise direction of movement)}

Because the sigmoid term $\sigma(\theta_t)
      \left(1 - \sigma(\theta_t)\right)$ is always non-negative,
the direction of change for $\theta_t$ during
gradient descent depends solely on the sign of $S_t$,

\[
      \operatorname{sign}\!\left(-\frac{\partial \mathcal{L}}
      {\partial \theta_t}\right)
      = \operatorname{sign}(S_t).
\]
Thus, If $S_t > 0$, the loss decreases when
$\theta_t$ increases, making $\sigma(\theta_t)$ larger.
In this case, the task shifts its weight toward the \emph{outlier} component.
Conversely, if $S_t < 0$, the loss decreases
when $\theta_t$ decreases, making $\sigma(\theta_t)$ smaller,
and the task shifts its weight toward the
\emph{non-outlier} component.
The sigmoid factor ensures smooth and bounded updates,
preventing any single task from dominating the optimization.
The exact direction (whether $\sigma \to 0$
corresponds to the \emph{non-outlier} or \emph{outlier}
component) depends on the initialization convention,
but the mechanism consistently drives $\theta_t$ toward
the component that better reduces the loss for task $t$.

Regarding model complexity, the theoretical and empirical analyses presented in~\ref{appendix_b}
indicate that the additional optimization introduced in Block~2 for task-specific parameters
does not substantially increase computational cost compared to standard multi-task boosting.
The model scales efficiently with the number of tasks while maintaining training stability.

\section{Experiments and Results}
\label{sec:experiments}
\begin{sloppypar}
      To conduct the experiments, a combination of physical
      computing resources and developed code was utilized to
      support both the training and evaluation of the
      proposed and state-of-the-art models.
      The model was developed using \verb?Python?
      (\verb?version 3.9?), and
      \verb?scikit-learn? (\verb?version 1.6?)
      \footnote{\href{https://github.com/scikit-learn/scikit-learn}
            {github.com/scikit-learn}}~\citep{pedregosa2011scikit}.
      For transparency and reproducibility,
      the complete code-base, with the preprocessed dataset,
      has been made publicly accessible through the
      associated GitHub
      repository
      \footnote{\href{https://github.com/GAA-UAM/R-MTGB}
            {github.com/GAA-UAM/R-MTGB}}.
\end{sloppypar}
To evaluate the proposed \ac{R-MTGB} model,
we conducted experiments alongside
state-of-the-art models. These include:
(1) a conventional multi-task \ac{GB} model (\ac{MTGB})~\citep{Emami2023};
(2) \ac{ST}-\ac{GB},
a standard \ac{GB} model trained independently for
each task;
(3) a data pooling approach
where a standard \ac{GB} model is trained
on data from all tasks combined (\ac{DP}-\ac{GB});
and (4) a \ac{TaF}-\ac{GB}, which is another data pooling approach
in which the input data is augmented using an extra input feature with a
one-hot encoding of the corresponding task identifier associated to each instance.
The core implementations of
\ac{GB}, \ac{TaF}-\ac{GB},
and \ac{DP}-\ac{GB} are based on the standard \ac{GB}
framework proposed in~\cite{Friedman2001}.
In real-world datasets,
the models were trained and evaluated by randomly splitting the data into
training and testing subsets using an 80:20 ratio.
To ensure the reliability and robustness of the results,
this process was repeated $100$ times.
For the synthetic datasets, $100$ distinct train/test datasets were
generated.
For both real-world and synthetic datasets,
we report the average performance across all tasks,
followed by the average results computed over all repetitions.

We compare our proposed method, \ac{R-MTGB},
primarily with \ac{GB} and \ac{MTGB},
as these methods represent the most relevant baselines to compare with.
\ac{GB} provides the natural point of reference since
our method is an extension of this framework,
and both \ac{ST}-\ac{GB} and
pooled-task (\ac{DP}-\ac{GB}, \ac{TaF}-\ac{GB}) are variants that allow
us to quantify the benefits of incorporating an \ac{MTL} paradigm.
\ac{MTGB} is the closest existing approach, as it explicitly
models shared and task-specific components within boosting.
Nevertheless, \ac{MTGB} lacks robustness to adversarial or outlier tasks.
Direct comparison with \ac{MTGB} therefore isolates the contribution of
our outlier-aware design. By contrast, methods such as
\ac{TS}-\ac{GB}~\citep{Mingcheng2021}
and FederBoost~\citep{Liu_Haizhou2024} address orthogonal
challenges—\ac{TS}-\ac{GB} focuses on modifying tree
splitting criteria, and FederBoost targets privacy,
preserving distributed learning—
making them less appropriate for direct comparison.

To ensure consistent reporting and to facilitate fair
comparisons of models performance,
hyperparameters were optimized via 5-fold cross-validation
within-training using a grid search.
The grid search was configured to optimize \ac{RMSE} for
regression models and accuracy for classification models.
This procedure was applied to both the synthetic
(see Subsection~\ref{sub_sec:synthetic_experiments})
and the real-world datasets (see Subsection~\ref{sub_sec:results}).
Notably, all compared methods can be viewed as
particular instances of \ac{R-MTGB} framework,
differentiated by the number of estimators to use in each different block.
This generalization allows \ac{R-MTGB} to flexibly encompass a wide
range of model configurations under a unified framework.
The ranges of hyperparameter values explored for each method
are summarized in Table~\ref{table:hparams}.
Hyperparameters not listed in the table were
set to their default values as defined in the
\verb?scikit-learn?  library. Additionally, decision stumps
were used as the base learner for all the studied models.
\begin{table}[ht]
      \caption{Hyperparameter grid reporting the number of base learners
            considered for each method.}
      \label{table:hparams}
      \centering
      \renewcommand{\arraystretch}{1.1}
      \small
      \begin{tabularx}{\textwidth}{p{2.5cm}XXX}
            \toprule
            \multirow{2}{*}{\textbf{Model}}
                                     & \multicolumn{3}{c}{\makecell{\textbf{No. of Base learners}}}                                           \\
            \cmidrule(lr){2-4}
                                     & \makecell{\textbf{1st}                                                                                 \\ \textbf{Block}}
                                     & \makecell{\textbf{2nd}                                                                                 \\ \textbf{Block}}
                                     & \makecell{\textbf{3rd}                                                                                 \\ \textbf{Block}} \\
            \midrule
            \textbf{\textsc{R-MTGB}} & $[0, 20, 30, 50]$                                            & $[20, 30, 50]$ & $[0, 20, 30, 50, 100]$ \\
            \textbf{\textsc{MTGB}}   & $[20, 30, 50]$                                               & --             & $[0, 20, 30, 50, 100]$ \\
            \textbf{\textsc{ST-GB}}  & --                                                           & --             & $[20, 30, 50, 100]$    \\
            \textbf{\textsc{DP-GB}}  & $[20, 30, 50, 100]$                                          & --             & --                     \\
            \textbf{\textsc{TaF-GB}} & $[20, 30, 50, 100]$                                          & --             & --                     \\
            \bottomrule
      \end{tabularx}
\end{table}

\subsection{Synthetic Experiments}
\label{sub_sec:synthetic_experiments}

First, we conducted a series of experiments using synthetic data
to validate the robustness of the developed \ac{R-MTGB} model
and to test our hypothesis prior to evaluation on real-world datasets.
These synthetic datasets were generated
using a combination
of shared ($\phi(\mathbf{x})$) and
task-specific ($\psi^{(t)}(\mathbf{x})$) functions. These
functions are generated randomly using a framework based on
Random Fourier Features~\citep{Rahimi2007}.
Considering a multi-task dataset \(\mathcal{D}_{t}\) as defined in
Eq.~\eqref{eq:mtl_dataset}
and,
\begin{align*}
      \mathbf{x}_{i,t} \sim \mathcal{U}([-1, 1]^{d}),
      \quad \forall i = 1, \ldots, N^{(t)},
\end{align*}
for a given input $\mathbf{x} \in \mathbb{R}^d$, the
function output is generated with
\begin{equation}
      \label{eq:random_fourier_features}
      \Psi(\mathbf{x}) = \sum_{i=1}^{N} \theta_i
      \sqrt{\frac{2 \alpha}{D}}
      \cos\left( \mathbf{w}_i^\top
      \frac{\mathbf{x}}{d_x} + b_i \right),
\end{equation}
where,
each $\mathbf{w}_i \sim \mathcal{N}(\mathbf{0}, \mathbf{I})$
denotes a random frequency vector, while $b_i \sim \mathcal{U}(0, 2\pi)$
is a scalar phase shift. The random feature is weighted by
$\theta_i \sim \mathcal{N}(0,1)$, and the scaling hyperparameter is
given by $\alpha$. The effective smoothness factor is defined as
$d_x = 0.5  \cdot d$, where $d$ is the input dimension.
Finally, $D$ denotes the number of random features, which is set to be equal to $500$.

By using this generation process,
the latent functions ($\phi(\mathbf{x})$ and
$\psi^{(t)}(\mathbf{x})$) are approximately and independently
sampled from a Gaussian process (GP)
prior, as the Random Fourier Feature representation provides
an explicit approximation of functions drawn from a stationary GP $\Psi(\cdot)$,
\begin{equation*}
      \phi(\mathbf{x}) \sim \Psi(\mathbf{x}),
      \qquad
      \psi^{(t)}(\mathbf{x}) \sim \Psi(\mathbf{x}).
\end{equation*}
(See~\cite{wilson2020efficiently} for further details).

For a set of $T=T_\text{non-out}+T_\text{out}$ tasks, $T_\text{non-out}$ \textit{non-outlier}
tasks can be generated by using a function
defined as a combination of a common
function and a task-specific function,
\begin{equation}
      \label{eq:task_function_combination}
      f^{(non-out)}_t(\mathbf{x}) =
      w \cdot \phi(\mathbf{x})
      + (1-w) \cdot \psi^{(t)}(\mathbf{x}),
\end{equation}
where, $w$ is the combination weight, for $t=1,\ldots,T_\text{non-out}$.
Similarly, $T_\text{out}$ \textit{outlier} tasks, on the other hand, are generated
by replacing the shared function $\phi$ with a different function
$\phi^{\text{out}}$ that is independently sampled,
\begin{equation}
      \label{eq:task_function_combination_outlier}
      f^{(out)}_t(\mathbf{x}) =
      w
      \cdot \phi^{\text{out}}(\mathbf{x}) +
      (1-w) \cdot \psi^{(t)}(\mathbf{x}),
\end{equation}
for $t = T_{\text{non-out}}+1,\ldots, T$.
For each instance $\mathbf{x}_{i,t}$, the target is generated by setting
the continuous value $y_{i,t}$ equal to the output of
the corresponding function, \emph{i.e.},
$f_t^{(\text{non-out})}(\mathbf{x}_{i,t})$
or $f_t^{(\text{out})}(\mathbf{x}_{i,t})$, in
regression. In binary classification, the
class label is obtained by applying
the sign function to the output of
$f_t^{(\text{non-out})}(\mathbf{x}_{i,t})$ or
$f_t^{(\text{out})}(\mathbf{x}_{i,t})$.

\begin{figure}[!htbp]
      \centering
      \includegraphics[width=0.7\linewidth]{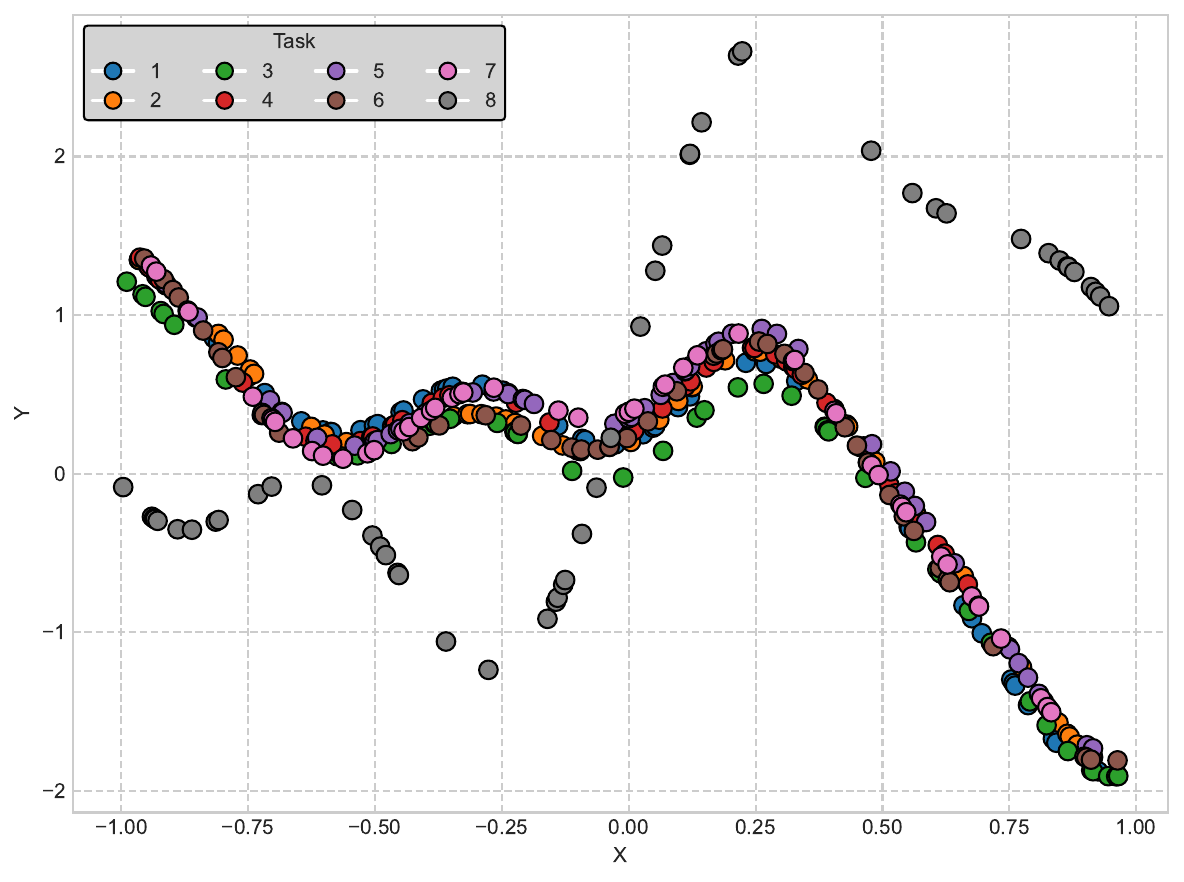}
      \caption{A visualization of
            the generated data points,
            comprising seven non-outlier (common)
            tasks (tasks 1 to 7) and one outlier
            task (task 8).
      }
      \label{fig:toy_data}
\end{figure}

An example of a generated toy dataset with one input and one
output dimension, comprising seven non-outlier (common) tasks
and one outlier task, is illustrated in Figure~\ref{fig:toy_data}.
The figure shows that the non-outlier tasks (tasks one to seven)
cluster together and
form a coherent band, indicating that they share an underlying
functional structure. In contrast,
task eight is distinguishable as an outlier.
The data points of task eight diverge significantly from the smooth
patterns observed in the other tasks.
This distinction arises because task eight was generated
using \(\phi^{\text{out}}\), which differs from \(\phi\), as described above.

Using the previously described techniques, we generated $100$
random batches of a synthetic toy dataset,
each initialized with a different random seed to
ensure diversity, with a weighting parameter of $w = 0.9$.
This guarantees different functions for each task, for each batch.
Each task consisted of $300$ training instances and $1{,}000$
test instances, distributed across five input features.
To preserve class balance, we ensured that each class
contained at least $10\%$ of the total samples.
Each batch included $10$ tasks in
total, two of which (last two tasks) were designated as outliers.
We consider regression and binary classification settings.
The generated datasets are publicly available on
Mendeley Data\footnote{\href{https://data.mendeley.com/datasets/r2mnkjfmh3}
      {Multi-Task synthetic dataset}}.
For each batch of experiments, the models
were trained using a fixed learning rate
of one and a decision stump as the base learner.
The number of base learners of each block was tuned using a 5-fold grid search
cross-validation method on the training set. For this, we considered the hyperparameter
grid defined in Table~\ref{table:hparams} for each method.
The best set of
hyperparameters found was then used for training
the model.
The performance of the evaluated models was
measured using recall and accuracy for classification
problem (Table \ref{tab:recall_accuracy}),
and \acf{RMSE} and \ac{MAE} for regression
problem (Table \ref{tab:mae_rmse}).

Table~\ref{tab:recall_accuracy} shows the results obtained in
the classification setting.
The best-performing method in each metric is indicated in \textbf{bold}.
The table shows that the introduced method, \ac{R-MTGB}, achieves
the highest test recall and accuracy
among all evaluated methods, indicating strong
generalization to unseen data.

\begin{table}[ht]
      \centering
      \footnotesize
      \caption{Average recall and accuracy scores with standard
            deviations, computed by first averaging across tasks
            and then over runs. Results are shown for each
            method on training and testing datasets, with best
            values per dataset in bold.}
      \begin{tabularx}{\textwidth}{lXXXX}
            \toprule
            \textbf{Model}           & \multicolumn{2}{c}{\textbf{Recall}} & \multicolumn{2}{c}{\textbf{Accuracy}}                                                           \\
                                     & \textbf{Train}                      & \textbf{Test}                         & \textbf{Train}             & \textbf{Test}              \\
            \midrule
            \textbf{\textsc{R-MTGB}} & 0.882 $\pm$ 0.090                   & \textbf{0.829 $\pm$ 0.108}            & 0.893 $\pm$ 0.046          & \textbf{0.843 $\pm$ 0.042} \\
            \textbf{\textsc{MTGB}}   & 0.895 $\pm$ 0.080                   & 0.824 $\pm$ 0.108                     & 0.905 $\pm$ 0.039          & 0.839 $\pm$ 0.042          \\
            \textbf{\textsc{DP-GB}}  & 0.773 $\pm$ 0.173                   & 0.755 $\pm$ 0.182                     & 0.794 $\pm$ 0.047          & 0.778 $\pm$ 0.049          \\
            \textbf{\textsc{ST-GB}}  & \textbf{0.901 $\pm$ 0.083}          & 0.819 $\pm$ 0.114                     & \textbf{0.911 $\pm$ 0.039} & 0.834 $\pm$ 0.043          \\
            \textbf{\textsc{TaF-GB}} & 0.801 $\pm$ 0.144                   & 0.782 $\pm$ 0.154                     & 0.816 $\pm$ 0.042          & 0.800 $\pm$ 0.045          \\
            \bottomrule
      \end{tabularx}
      \label{tab:recall_accuracy}
\end{table}

Regarding the regression setting,
Table~\ref{tab:mae_rmse} shows the results obtained.
Again, the best-performing methods per column are
indicated in \textbf{bold}.
We observe that the proposed
method, \ac{R-MTGB}, achieved the lowest \ac{MAE}
and \ac{RMSE} on the test set, indicating the most
accurate and robust regression performance on unseen data.
As in the classification problem, \ac{ST}-\ac{GB}
achieved slightly better performance on the
training set, but exhibited higher test errors
compared to \ac{R-MTGB}.
By contrast, the performance of \ac{R-MTGB} remained
consistent across training and testing for both problems,
highlighting its ability to generalize effectively to
unseen data.
The performance gaps between the training and test results
presented in Tables~\ref{tab:recall_accuracy} and~\ref{tab:mae_rmse}
for the different methods serve as indicators
of each method's regularization capability.
The proposed \ac{R-MTGB} method exhibits the smallest drop
in performance between the training and test sets,
demonstrating the most effective regularization among the
evaluated methods.
\begin{table}[ht]
      \centering
      \footnotesize
      \caption{Average \ac{MAE} and \ac{RMSE} scores with standard
            deviations, computed by first averaging across tasks
            and then over runs. Results are shown for each
            method on training and testing datasets, with best
            values per dataset in bold.}
      \begin{tabularx}{\textwidth}{lXXXX}
            \toprule
            \textbf{Model}           & \multicolumn{2}{c}{\textbf{MAE}} & \multicolumn{2}{c}{\textbf{RMSE}}                                                           \\
                                     & \textbf{Train}                   & \textbf{Test}                     & \textbf{Train}             & \textbf{Test}              \\
            \midrule
            \textbf{\textsc{R-MTGB}} & 0.309 $\pm$ 0.041                & \textbf{0.332 $\pm$ 0.043}        & 0.397 $\pm$ 0.053          & \textbf{0.426 $\pm$ 0.055} \\
            \textbf{\textsc{MTGB}}   & 0.265 $\pm$ 0.038                & 0.359 $\pm$ 0.048                 & 0.340 $\pm$ 0.049          & 0.466 $\pm$ 0.061          \\
            \textbf{\textsc{DP-GB}}  & 0.444 $\pm$ 0.089                & 0.470 $\pm$ 0.091                 & 0.583 $\pm$ 0.121          & 0.617 $\pm$ 0.125          \\
            \textbf{\textsc{ST-GB}}  & \textbf{0.260 $\pm$ 0.044 }      & 0.365 $\pm$ 0.048                 & \textbf{0.333 $\pm$ 0.056} & 0.472 $\pm$ 0.062          \\
            \textbf{\textsc{TaF-GB}} & 0.387 $\pm$ 0.062                & 0.412 $\pm$ 0.065                 & 0.504 $\pm$ 0.081          & 0.537 $\pm$ 0.085          \\
            \bottomrule
      \end{tabularx}
      \label{tab:mae_rmse}
\end{table}

\begin{figure}[!htbp]
      \centering
      \includegraphics[width=\linewidth]{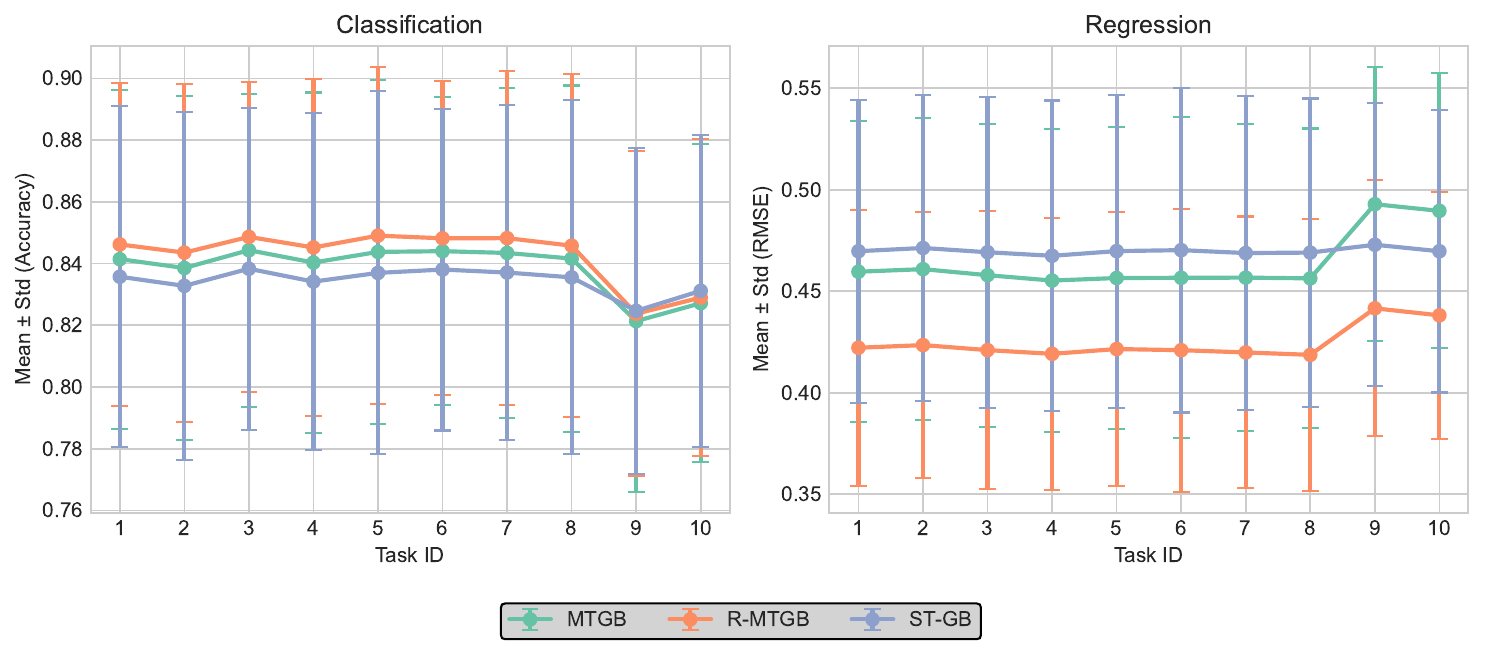}
      \caption{Average task-wise performance of the
            evaluated models over multiple runs shown
            separately for classification (left subplot)
            and regression (right subplot) tasks.}
      \label{fig:performance_across_tasks_synthetic_data}
\end{figure}

Figure~\ref{fig:performance_across_tasks_synthetic_data} presents
the average performance and standard deviation of each task-wise model,
evaluated on the unseen portion of the same synthetic
dataset described previously.
The results are reported separately for classification tasks,
using accuracy (left subplot), and for regression tasks,
using \ac{RMSE} (right subplot).
Each model is depicted using a distinct color, with vertical lines
around the means
representing the standard deviation.
From Figure~\ref{fig:performance_across_tasks_synthetic_data},
it can be observed that
the proposed \ac{R-MTGB} model
outperforms both \ac{MTGB} and \ac{ST}-\ac{GB} in all regression
and most classification tasks.
For classification
(Figure~\ref{fig:performance_across_tasks_synthetic_data}, left subplot),
\ac{R-MTGB} model achieves the highest
mean accuracy in the common
tasks. For the outlier tasks, the studied
models demonstrate
comparable performance. Here, \ac{R-MTGB}
surpasses \ac{MTGB},
whereas \ac{ST}-\ac{GB}
performs negligibly better.
Moreover,
in the regression tasks
(Figure~\ref{fig:performance_across_tasks_synthetic_data},
right subplot),
\ac{R-MTGB} outperforms all methods, in outlier and non-outlier tasks,
demonstrating
robustness and effective utilization of information from common
tasks to achieve a strong performance across all tasks.
Importantly, the absence of a weighting mechanism to
identify outliers, causes \ac{MTGB}
approach to underperform in outlier tasks across both
problem settings.
Additional experiments, including the
training and evaluation of a \ac{DNN} model,
are presented in~\ref{appendix_a}, showing that
while \ac{DNN} performed reasonably well on common tasks,
it was less effective on outlier tasks, confirming the
superior robustness and generalization of the proposed \ac{R-MTGB}
model.

\begin{figure}[!htbp]
      \centering
      \includegraphics[width=0.7\linewidth]{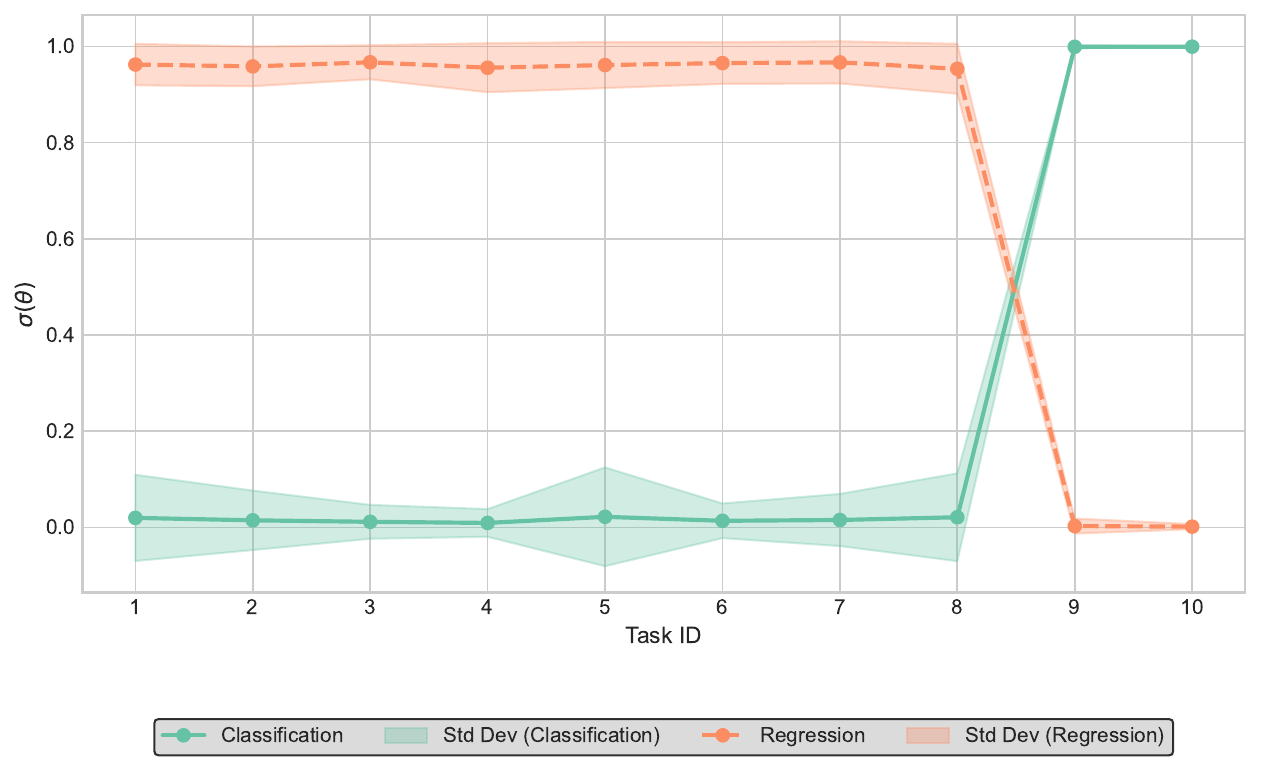}
      \caption{Mean and standard deviation of
            \(\sigma(\boldsymbol{\theta})\) for
            each task learned by \ac{R-MTGB} model on
            the generated synthetic multi-task data.
            Values of \(\sigma(\boldsymbol{\theta})\) near 0
            or 1 indicate task separation, with one
            extreme representing non-outlier tasks and
            the opposite extreme representing outlier tasks;
            the specific direction
            (0 = non-outlier vs. 1 = outlier, or vice versa)
            may depend on the problem.
      }
      \label{fig:sigmoid_thetas_synthetic}
\end{figure}

To assess the ability of the presented model to detect and
identify the two defined outlier tasks,
the mean and standard deviation of the learned
\(\sigma(\boldsymbol{\theta})\)
values by \ac{R-MTGB} model are visualized in
Figure~\ref{fig:sigmoid_thetas_synthetic} for regression
and classification problems.
The lines depict the mean values, while the shaded areas
represent the corresponding standard deviations across multiple
batches of experiments.
The results for the classification setting are shown as a
solid line with light green shading,
whereas the results for regression are depicted with a
dashed line and light orange shading.
The \(\sigma(\boldsymbol{\theta})\) values should
split tasks into outliers and
non-outliers, but it is not clear which extreme values
will be assigned to outlier or non-outlier tasks due
to the random initialization of
the \(\boldsymbol{\theta}\)
values (See Subsection~\ref{subsec:theoretical_analysis},
\textit{Proposition 1}, for further details).
In any case,
the results demonstrate that \ac{R-MTGB}
effectively identified outlier tasks by
assigning them weights at the opposite extremes
compared to non-outlier tasks.
The small standard deviation observed across tasks,
especially for the outlier tasks, further highlights the
robustness of the designed parameter optimization.

\subsubsection{Key Benefits of RMTGB in Estimating Shared Functions}
\label{sub_sec:illustrative_experiment}

\begin{figure}[!htbp]
      \centering
      \includegraphics[width=0.8\linewidth]{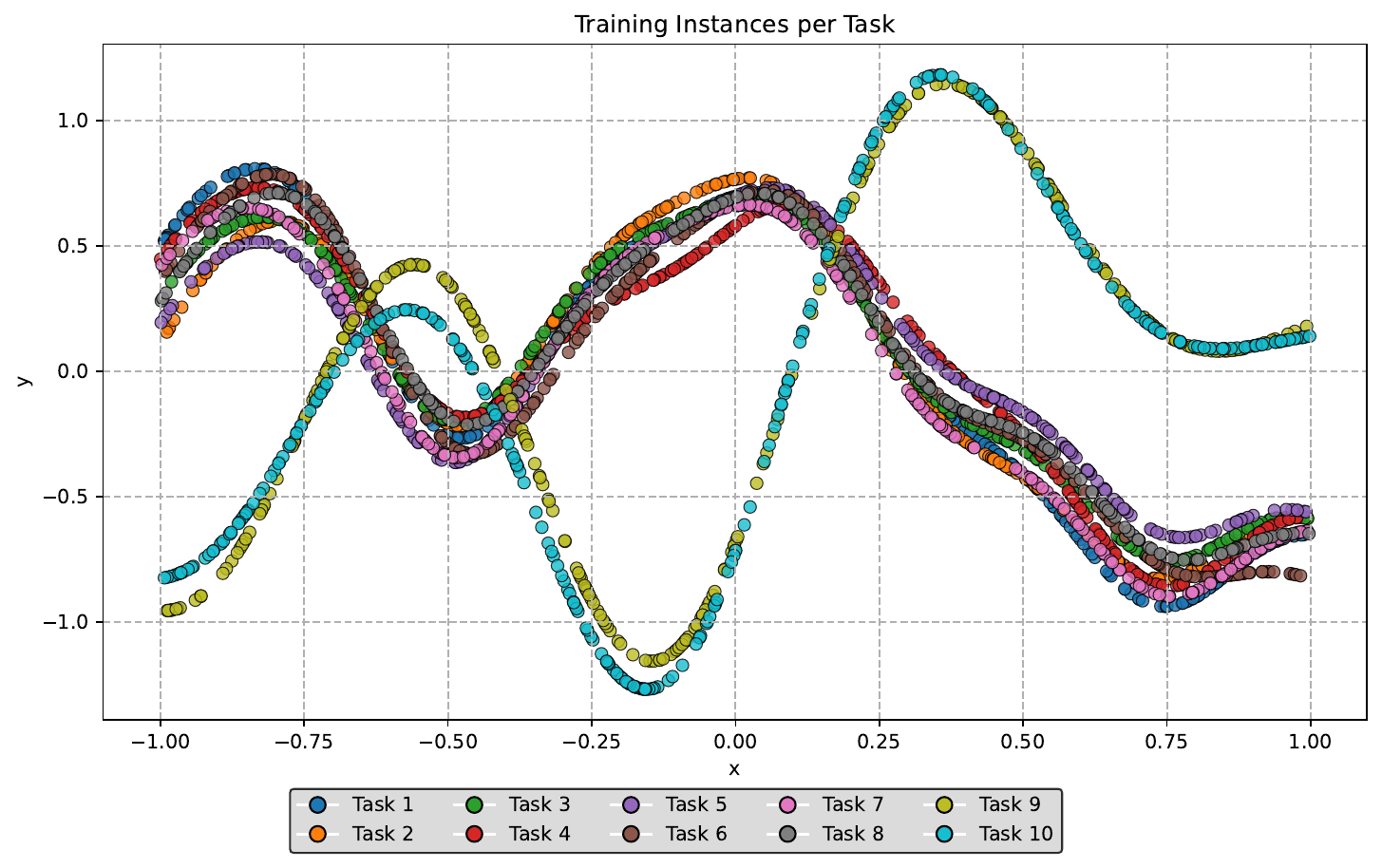}
      \caption{A visualization of the distribution of
            training data points,
            comprising eight non-outlier
            tasks (Tasks 1-8) and two outlier
            tasks (Tasks 9-10).}
      \label{fig:training_instancese}
\end{figure}

To further examine the effect of outlier
tasks on shared function estimation in \ac{MTL}
models, we conducted an additional experiment using a
one-dimensional toy dataset generated
by Eq.~\eqref{eq:random_fourier_features}.
The dataset consisted of 10 tasks, including eight
non-outlier tasks and two outlier tasks. Each
task had 300 training instances, instantiated
with both a shared component and a
task-specific component (see Eq.~\eqref{eq:shared_additive}
for further details).
Figure~\ref{fig:training_instancese}
illustrates the generated training data across
all tasks, with tasks one to eight categorized
as non-outlier tasks, and tasks nine and ten
designated as outlier tasks.

The purpose of this experiment is to see
the effect of outlier tasks in the estimation of the common
or shared function assumed by each method \ac{MTGB} and \ac{R-MTGB}.
Specifically, it is expected that outlier tasks severely impair the
estimation process in \ac{MTGB}. Recall that \ac{MTGB} assumes all tasks have a
shared common function.  To test this, we trained both \ac{MTGB} and \ac{R-MTGB}
on this dataset using 150 base learners. Namely, 150 shared base learners for \ac{MTGB}
and 50 shared base learners (Block 1) and
100 outlier-aware base learners (Block 2), in \ac{R-MTGB}. We did not consider any
task-specific fitting, \emph{i.e.}, the number of base learners
in Block 3 is set equal to $0$ in both \ac{MTGB} and \ac{R-MTGB}.
We plotted the estimated functions obtained by each method across tasks.
These functions should try to fit $\phi(\mathbf{x})$
and $\phi^{\text{out}}(\mathbf{x})
$ in Eq. (\ref{eq:task_function_combination}) and
Eq. (\ref{eq:task_function_combination_outlier}), respectively.
Furthermore, we compare the corresponding estimates against
the ground-truth function shared among tasks using in the data generation process.

\begin{figure}[!htbp]
      \centering
      \includegraphics[width=1.0\linewidth]{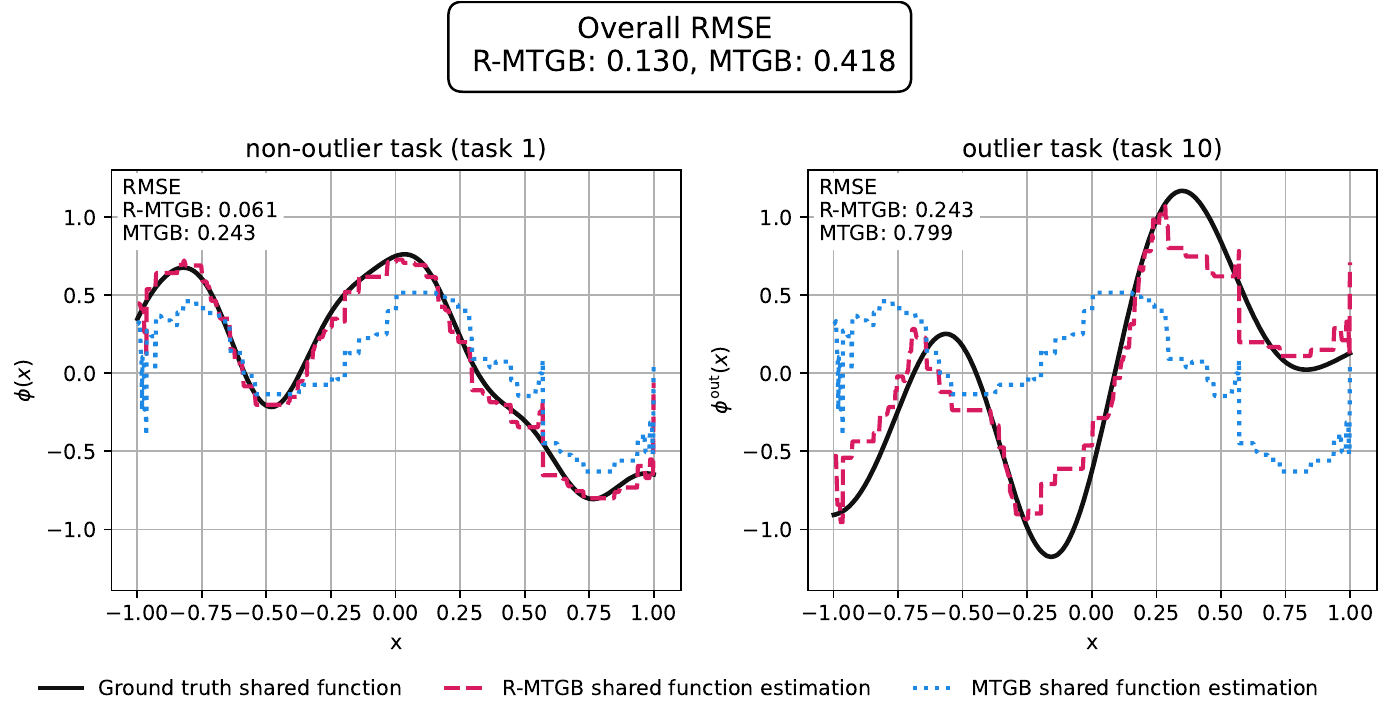}
      \caption{Comparison of shared function estimation results by \ac{R-MTGB} and \ac{MTGB} for a
            representative non-outlier task (left subplot) and a representative outlier task (right subplot).}
      \label{fig:shared_function_example}
\end{figure}

Figure~\ref{fig:shared_function_example} shows the results for
task one (non-outlier) and task ten (outlier)
as representative examples, since the remaining tasks exhibit
similar behavior.
For each method, the figure shows
performance in terms of \ac{RMSE} estimation with respect to the
actual function, \emph{i.e.},
$\phi(\mathbf{x})$ or $\phi^{\text{out}}(\mathbf{x})$.
Overall generalization performance
across all tasks is displayed in the figure title.
We observe that the second block of the \ac{R-MTGB} model enables
it to correctly approximate the ground-truth shared function
for both non-outlier task (left subplot) and outlier task
(right subplot). By contrast, \ac{MTGB} enforces a
single shared function across all tasks,
which becomes biased by the presence of
outlier tasks, leading to poor fit for both components.
In this experiment, the third block was not used
(zero iterations), so the improvements stem
solely from the combined effect of Block 1
(initial global shared-learning) and Block 2
(outlier-aware task partitioning).
This behavior of \ac{R-MTGB} prevents distortion of the
shared representation, unlike in \ac{MTGB}, ensuring that
non-outlier tasks retain an accurate shared component,
while outlier tasks are modeled jointly through a
separate shared component.
As a result, \ac{R-MTGB} achieves substantially
lower overall error compared to \ac{MTGB}, demonstrating
its robustness in estimating shared
functions under task heterogeneity.
A better estimation of the shared component is expected to result in better
generalization performance in real-world problems.

\subsection{Real-World Datasets Results}
\label{sub_sec:results}

Table~\ref{table:dataset} presents a summary of
the real-world datasets considered in our experiments,
including their references, number of instances, features,
tasks, and field of application. They are divided into two categories:
classification and regression. Among the classification
datasets, the \textit{Avila} dataset is a multi-class
problem with 12 classes, while the others are binary
classification datasets.
The distribution of instances across classes shows different
levels of imbalance:
in the \textit{Adult} dataset, class~0 includes 37,155
instances and class~1 includes 11,687;
in \textit{Landmine}, there are 13,916 instances of
class~0 and only 904 of class~1;
and in \textit{Bank Marketing}, class~0 has 39,922
instances compared to 5,289 of class~1.
The \textit{Avila} dataset is the most imbalanced,
with class sizes ranging from just 10 instances (class~11)
to 8,572 (class~0).
In all the studied models, each sample contributes
equally to the cross-entropy
loss function (Eq.~\eqref{eq:cross_entropy}),
and the boosting process minimizes the average
loss across all samples.
Tasks are defined according to
the natural structure of each dataset,
such as copyist attribution (\textit{Avila}), demographic groups
(\textit{Adult}),
occupational categories (\textit{Bank Marketing}),
or sensor fields (\textit{Landmine}). Similarly,
the regression datasets are organized into tasks based on
intrinsic attributes of the data, such as biological
categories (\textit{Abalone}),
individual participants (\textit{Parkinsons}), robotic joints,
(\textit{SARCOS}), school identifiers (\textit{School}),
or participant groups (\textit{Computer}).
These datasets have been widely
adopted as benchmarks in recent \ac{MTL} studies for
both classification~\citep{Zhao2018,Oneto2019,Wang2021,Emami2023}
and regression~\citep{Argyriou2007,Argyriou2008,Ciliberto2017,
      Gunduz2019,Wang2022,Srinivasan2024} problems.
\begin{table}[!htb]
      \caption{Real-world datasets description.}
      \label{table:dataset}
      \resizebox{\textwidth}{!}{%
            \centering
            \begin{tabular}{lllll}
                  \toprule
                  \textbf{Name}                   & \textbf{Instances} & \textbf{Attributes} & \textbf{Tasks} & \textbf{Field}          \\
                  \midrule
                  \multicolumn{5}{l}{\textbf{Classification}}                                                                           \\
                  Avila~\citep{Stefano2018}       & $20,867$           & $10$                & $48$           & Handwriting Recognition \\
                  Adult~\citep{Becker1996}        & $48,842$           & $14$                & $7$            & Social Science          \\
                  Bank Marketing~\citep{Moro2011} & $45,211$           & $16$                & $12$           & Marketing               \\
                  Landmine~\citep{Yilmaz2018}     & $14,820$           & $9$                 & $29$           & Engineering             \\
                  \midrule
                  \multicolumn{5}{l}{\textbf{Regression}}                                                                               \\
                  Abalone~\citep{Nash1994}        & $4,177$            & $8$                 & $3$            & Bioinformatics          \\
                  Computer~\citep{Lenk1996}       & $3,800$            & $13$                & $190$          & Survey                  \\
                  Parkinsons~\citep{Tsanas2009}   & $5,875$            & $19$                & $42$           & Biomedical              \\
                  SARCOS~\citep{Jawanpuria2015}   & $342,531$          & $21$                & $7$            & Robotics                \\
                  School~\citep{Bakker2003}       & $15,362$           & $10$                & $139$          & Social Science          \\
                  \bottomrule
            \end{tabular}
      }
\end{table}

The experimental results of the evaluated
models on real-world datasets
(listed in Table~\ref{table:dataset}) are
summarized in Tables~\ref{tab:accuracy} through~\ref{tab:mae}.
All reported metrics are computed by first averaging
across all tasks within each batch,
and then calculating the mean and standard deviation
over 100 batch runs. Note that each dataset contains a
different number of tasks.

Specifically, Tables~\ref{tab:accuracy}
and~\ref{tab:recall} present the average
accuracy and unweighted mean recall on unseen test
data for the classification datasets.
Similarly, Tables~\ref{tab:rmse} and~\ref{tab:mae}
report the average \ac{RMSE} and \ac{MAE} on the test
sets for the regression datasets.
The best-performing results in each category
are indicated in \textbf{boldface} per column.

The results in Table~\ref{tab:accuracy} clearly show
that the proposed \ac{R-MTGB} model consistently achieves
the highest testing accuracy on all datasets, either matching or
exceeding the performance of the competing methods.
Notably, \ac{R-MTGB} outperforms the other models on four
out of five datasets and ties with \ac{MTGB} on
the \textit{Landmine} dataset.

\begin{table}[ht]
      \centering
      \footnotesize
      \caption{Testing accuracy across models for each dataset,
            averaged first over tasks within each batch
            and then over runs. Mean and standard deviation
            are reported, with best values per dataset shown in bold.}
      \setlength{\extrarowheight}{2pt}
      \begin{tabularx}{\textwidth}{l
                  >{\centering\arraybackslash}X
                  >{\centering\arraybackslash}X
                  >{\centering\arraybackslash}X
                  >{\centering\arraybackslash}X
                  >{\centering\arraybackslash}X}
            \toprule
            \textbf{Model}           & \textbf{Adult (Gender)}  & \textbf{Adult (Race) }   & \textbf{Avila}           & \textbf{Bank Marketing}  & \textbf{Landmine}        \\
            \midrule
            \textbf{\textsc{R-MTGB}} & \textbf{0.8493 ± 0.0036} & \textbf{0.8487 ± 0.0036} & \textbf{0.6190 ± 0.0465} & \textbf{0.8947 ± 0.0031} & \textbf{0.9428 ± 0.0035} \\
            \textbf{\textsc{MTGB}}   & 0.8479 ± 0.0036          & 0.8451 ± 0.0036          & 0.6138 ± 0.0503          & 0.8934 ± 0.0030          & \textbf{0.9428 ± 0.0035} \\
            \textbf{\textsc{DP-GB}}  & 0.8368 ± 0.0046          & 0.8368 ± 0.0046          & 0.4939 ± 0.0095          & 0.8889 ± 0.0029          & 0.9387 ± 0.0035          \\
            \textbf{\textsc{ST-GB}}  & 0.8406 ± 0.0036          & 0.8385 ± 0.0036          & 0.6099 ± 0.0605          & 0.8917 ± 0.0030          & 0.9423 ± 0.0035          \\
            \textbf{\textsc{TaF-GB}} & 0.8368 ± 0.0046          & 0.8368 ± 0.0046          & 0.4970 ± 0.0090          & 0.8889 ± 0.0029          & 0.9387 ± 0.0035          \\
            \bottomrule
      \end{tabularx}
      \label{tab:accuracy}
\end{table}

As shown in Table~\ref{tab:recall}, \ac{R-MTGB}
achieves the highest recall on three out of the five
datasets:
\textit{Adult (Gender)}, \textit{Adult (Race)},
and \textit{Bank Marketing}, indicating strong overall
performance across diverse classification tasks.
\ac{ST}-\ac{GB} slightly outperforms \ac{R-MTGB} on \textit{Avila}
and delivers the best
result on \textit{Landmine}, although the margin is
small. In contrast, pooling-based approaches exhibit
consistently lower recall and accuracy across all datasets,
particularly on complex datasets such as \textit{Avila}.
Overall, these results further validate the
effectiveness of \ac{R-MTGB} in leveraging
relational structure to enhance predictive performance across tasks.

An additional set of experimental results in
terms of the F1 score is provided in~\ref{appendix_a}.
These results show patterns consistent with the
accuracy and recall findings: \ac{R-MTGB} generally achieves the
highest or near-highest F1 scores across most
datasets, particularly in \textit{Adult (Gender)},
\textit{Adult (Race)}, and \textit{Bank Marketing},
while \ac{ST}-\ac{GB} slightly outperforms it on
\textit{Avila} and \textit{Landmine}. Overall, the
F1-score analysis confirms that the proposed model
maintains strong and balanced classification performance
across heterogeneous tasks.

\begin{table}[ht]
      \centering
      \footnotesize
      \caption{Testing recall across models for each dataset,
            averaged first over tasks within each batch
            and then over runs. Mean and standard deviation
            are reported, with best values per dataset shown in bold.}
      \setlength{\extrarowheight}{2pt}
      \begin{tabularx}{\textwidth}{l
                  >{\centering\arraybackslash}X
                  >{\centering\arraybackslash}X
                  >{\centering\arraybackslash}X
                  >{\centering\arraybackslash}X
                  >{\centering\arraybackslash}X}
            \toprule
            \textbf{Model}           & \textbf{Adult (Gender)}  & \textbf{Adult (Race) }   & \textbf{Avila}           & \textbf{Bank Marketing}  & \textbf{Landmine}        \\
            \midrule
            \textbf{\textsc{R-MTGB}} & \textbf{0.7256 ± 0.0057} & \textbf{0.7248 ± 0.0052} & 0.4360 ± 0.0832          & \textbf{0.5892 ± 0.0094} & 0.5478 ± 0.0099          \\
            \textbf{\textsc{MTGB}}   & 0.7205 ± 0.0051          & 0.7158 ± 0.0080          & 0.4372 ± 0.0853          & 0.5765 ± 0.0099          & 0.5485 ± 0.0100          \\
            \textbf{\textsc{DP-GB}}  & 0.6838 ± 0.0099          & 0.6838 ± 0.0099          & 0.1660 ± 0.0083          & 0.5321 ± 0.0037          & 0.5 ± 0.0000             \\
            \textbf{\textsc{ST-GB}}  & 0.7049 ± 0.0056          & 0.6931 ± 0.0083          & \textbf{0.4534 ± 0.0881} & 0.5552 ± 0.0050          & \textbf{0.5518 ± 0.0097} \\
            \textbf{\textsc{TaF-GB}} & 0.6838 ± 0.0099          & 0.6838 ± 0.0099          & 0.1689 ± 0.0069          & 0.5321 ± 0.0037          & 0.5 ± 0.0000             \\
            \bottomrule
      \end{tabularx}
      \label{tab:recall}
\end{table}

Regarding the regression datasets
and \ac{RMSE} metric (Table~\ref{tab:rmse}),
\ac{R-MTGB} achieves the lowest test errors on
nearly all datasets,
demonstrating remarkable effectiveness,
especially on datasets with structurally
complex tasks
(e.g., \textit{SARCOS}).
\ac{ST}-\ac{GB}, while not the top performer overall,
achieves the best results on the \textit{Parkinsons} dataset
with a small margin over \ac{R-MTGB}.
pooling-based methods like \ac{DP}-\ac{GB}
generally underperform, especially on more
complex datasets like \textit{Parkinsons} and \textit{SARCOS}.
\begin{table}[ht]
      \centering
      \footnotesize
      \caption{Testing \ac{RMSE} across models for each dataset,
            averaged first over tasks within each batch
            and then over runs. Mean and standard deviation
            are reported, with best values per dataset shown in bold.}
      \setlength{\extrarowheight}{2pt}
      \begin{tabularx}{\textwidth}{l
                  >{\centering\arraybackslash}X
                  >{\centering\arraybackslash}X
                  >{\centering\arraybackslash}X
                  >{\centering\arraybackslash}X
                  >{\centering\arraybackslash}X}
            \toprule
            \textbf{Model}           & \textbf{Abalone}         & \textbf{Computer}        & \textbf{Parkinson}       & \textbf{SARCOS}          & \textbf{School}           \\
            \midrule
            \textbf{\textsc{R-MTGB}} & \textbf{2.2660 ± 0.0857} & \textbf{2.4632 ± 0.0706} & 0.2868 ± 0.0316          & \textbf{4.7031 ± 0.0729} & \textbf{10.1313 ± 0.1262} \\
            \textbf{\textsc{MTGB}}   & 2.2894 ± 0.0866          & 2.4856 ± 0.0473          & 0.3355 ± 0.0243          & 4.8083 ± 0.0336          & 10.1536 ± 0.1222          \\
            \textbf{\textsc{DP-GB} } & 2.3970 ± 0.0935          & 2.4658 ± 0.0478          & 8.8586 ± 0.1373          & 18.3971 ± 0.0669         & 10.4229 ± 0.1176          \\
            \textbf{\textsc{ST-GB} } & 2.3464 ± 0.0889          & 2.7596 ± 0.3516          & \textbf{0.2684 ± 0.0274} & 4.9193 ± 0.0340          & 10.2952 ± 0.1366          \\
            \textbf{\textsc{TaF-GB}} & 2.3830 ± 0.0926          & 2.4668 ± 0.0669          & 6.5588 ± 0.0871          & 11.2658 ± 0.0597         & 10.4152 ± 0.1166          \\
            \bottomrule
      \end{tabularx}
      \label{tab:rmse}
\end{table}
In terms of \ac{MAE} results (Table~\ref{tab:mae}),
\ac{R-MTGB} again demonstrates consistent superiority,
achieving the lowest errors
on the same datasets as those on \ac{RMSE},
as shown in
Table~\ref{tab:rmse}.
\ac{ST}-\ac{GB} achieves the lowest \ac{MAE}
on \textit{Parkinsons},
again with a minor difference
compared to \ac{R-MTGB}.
As with the \ac{RMSE} results,
pooling-based methods such as \ac{DP}-\ac{GB} lag behind,
particularly on complex datasets like \textit{Parkinsons}
and \textit{SARCOS}.
These results confirm the overall trend that \ac{R-MTGB}
offers a
competitive and efficient alternative, striking a balance
between accuracy and scalability by capturing relational
information across tasks in a single model.
\begin{table}[ht]
      \centering
      \footnotesize
      \caption{Testing \ac{MAE} across models for each dataset,
            averaged first over tasks within each batch
            and then over runs. Mean and standard deviation
            are reported, with best values per dataset shown in bold.}
      \setlength{\extrarowheight}{2pt}
      \begin{tabularx}{\textwidth}{l
                  >{\centering\arraybackslash}X
                  >{\centering\arraybackslash}X
                  >{\centering\arraybackslash}X
                  >{\centering\arraybackslash}X
                  >{\centering\arraybackslash}X}
            \toprule
            \textbf{Model}            & \textbf{Abalone}         & \textbf{Computer}        & \textbf{Parkinson}       & \textbf{SARCOS}          & \textbf{School}          \\
            \midrule
            \textbf{\textsc{R-MTGB}}  & \textbf{1.6073 ± 0.0459} & \textbf{2.0208 ± 0.0610} & 0.1315 ± 0.0253          & \textbf{2.7366 ± 0.0343} & \textbf{8.0048 ± 0.1018} \\
            \textbf{\textsc{MTGB}}    & 1.6236 ± 0.0468          & 2.0536 ± 0.0437          & 0.1858 ± 0.0154          & 2.7778 ± 0.0156          & 8.0314 ± 0.0955          \\
            \textbf{\textsc{DP-GB}}   & 1.7322 ± 0.0502          & 2.0249 ± 0.0432          & 7.3403 ± 0.1180          & 12.6503 ± 0.0469         & 8.2701 ± 0.0975          \\
            \textbf{\textsc{ST-GB}}   & 1.6643 ± 0.0481          & 2.1966 ± 0.3169          & \textbf{0.1099 ± 0.0082} & 2.7783 ± 0.0154          & 8.1460 ± 0.1083          \\
            \textbf{\textsc{TaF-GB} } & 1.7110 ± 0.0502          & 2.0430 ± 0.0556          & 5.7127 ± 0.0882          & 7.1316 ± 0.0305          & 8.2696 ± 0.0975          \\
            \bottomrule
      \end{tabularx}
      \label{tab:mae}
\end{table}
In comparison with
\ac{MTGB} model, the proposed \ac{R-MTGB}
consistently outperforms it across all metrics,
tasks, and datasets. This demonstrates that
the proposed approach effectively addresses
the limitations of \ac{MTGB}
by incorporating a dynamic task weighting
mechanism.

\begin{figure}[!htbp]
      \centering
      \includegraphics[width=1.0\linewidth]{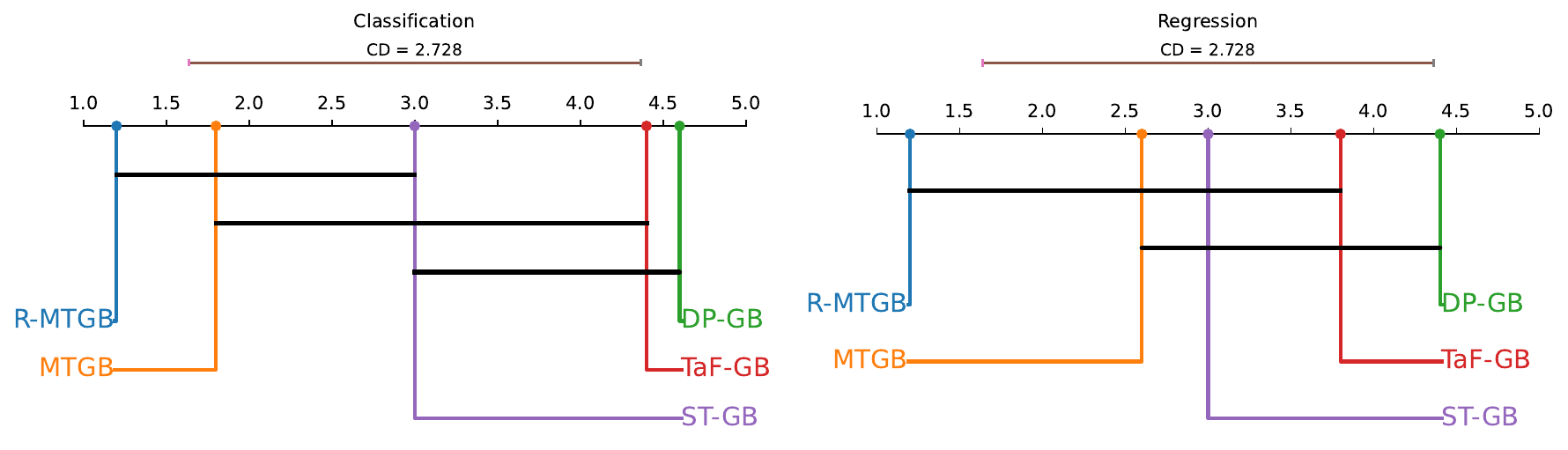}
      \caption{Dems\v{a}r plots with the Nemenyi test
            ($p=0.05$) comparing model performance across datasets.
            Colors follow model rank order.
            The x-axis shows average ranks over 100 runs
            (lower is better). Horizontal bars mark no
            significant difference; the \ac{CD} is shown at the top.}
      \label{fig:demsar_data}
\end{figure}

\begin{figure}[!htbp]
      \centering
      \includegraphics[width=1.0\linewidth]{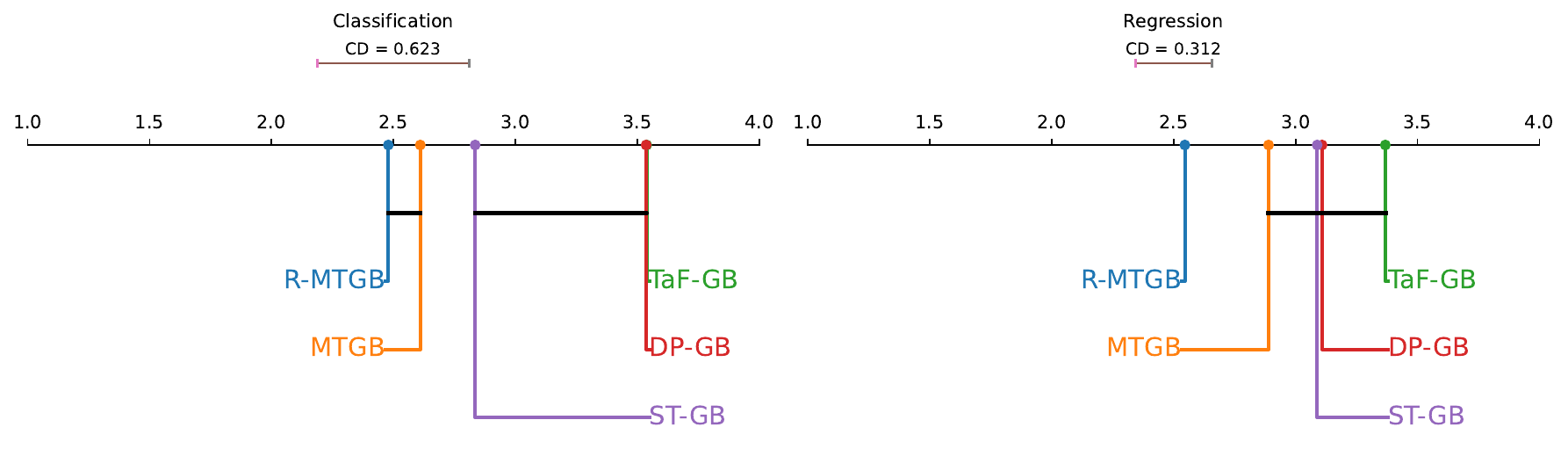}
      \caption{Dems\v{a}r plots with the Nemenyi test
            ($p=0.05$) comparing model performance across tasks.
            Colors follow model rank order.
            The x-axis shows average ranks over 100 runs
            (lower is better). Horizontal bars mark no
            significant difference; the \ac{CD} is shown at the top.}
      \label{fig:demsar_task}
\end{figure}

To systematically compare model performance across datasets
(dataset-wise) and tasks (task-wise),
as shown in Figures~\ref{fig:demsar_data} and~\ref{fig:demsar_task},
we employ Dems\v{a}r plots alongside the Nemenyi post-hoc
test~\citep{demvsar2006statistical}, using a significance level
of $p = 0.05$.
A Dem\v{s}ar plot shows the average rank of each model across
all evaluation scenarios.
In the dataset-wise scenario, models are evaluated on each task,
and performance is first averaged across all tasks and then
averaged over 100 repetitions. In the task-wise scenario,
the performance of each model on each task is averaged
over 100 repetitions.
In both scenarios, models are then ranked according to
their averaged performance:
in descending order for classification
(where higher accuracy is better) and in ascending order
for regression (where lower \ac{RMSE} is better).
The best-performing model receives rank one, the second-best
rank two, and so on.
The Dems\v{a}r plot
places each model along the horizontal axis according
to this average rank, where models closer to the left have lower
(better) ranks, indicating stronger overall performance.
Finally, to determine whether differences between models
are statistically significant, we apply the Nemenyi post-hoc test,
which calculates \ac{CD}. If the average ranks of two models
differ by more than the \ac{CD}, their performance difference is
considered statistically significant. In the corresponding plots,
this is shown with horizontal bars: models connected by
a bar are not significantly different in performance,
and the calculated \ac{CD}s are indicated above each subplot.

Our dataset-wise evaluation, as shown in
Figure~\ref{fig:demsar_data}, covers ten datasets in total:
five classification datasets (left subplot)
and five regression datasets (right subplot).
Figure~\ref{fig:demsar_data} shows that
\ac{R-MTGB} model achieves the lowest (best) average
rank for both classification and regression problems.
Notably, \ac{R-MTGB} maintains the best rank,
followed by \ac{MTGB}, with a larger
margin for regression
(Figure~\ref{fig:demsar_data}, right subplot).
The consistently poor performance of \ac{DP}-\ac{GB} and \ac{TaF}-\ac{GB}
across all scenarios in
Figures~\ref{fig:demsar_data}
indicates that simple data aggregation can
harm model effectiveness, likely due to the loss of
task-specific distinctions.
In contrast, the regularization mechanism in \ac{R-MTGB}
effectively leverages beneficial inter-task relationships
while mitigating the negative effects of unrelated
tasks, which is especially valuable in heterogeneous
task environments.

For a more granular view, we perform a task-wise comparison
by applying the same statistical procedure
to individual tasks (Figure~\ref{fig:demsar_task}).
For the ranking, the performance of each model on
each task is averaged across all repetitions.
This analysis includes $96$ classification
tasks measured by accuracy
(left subplot) and $381$ regression tasks,
with performance measured by \ac{RMSE} (right subplot).
Figure~\ref{fig:demsar_task}, left subplot, shows that \ac{R-MTGB}
model achieves the lowest (best) average rank.
Moreover, there is no statistically significant difference
between the proposed model and \ac{MTGB}; however,
both methods outperform the remaining models, with
the differences being statistically significant
relative to the other evaluated approaches.
In the regression tasks shown in the right subplot,
the introduced model once again achieves the best
(lowest) average rank. This improvement is
statistically significant compared to all other evaluated
methods, while the remaining models form a single
group with no significant differences among them.

\begin{figure}[!htbp]
      \centering
      \includegraphics[width=\linewidth]{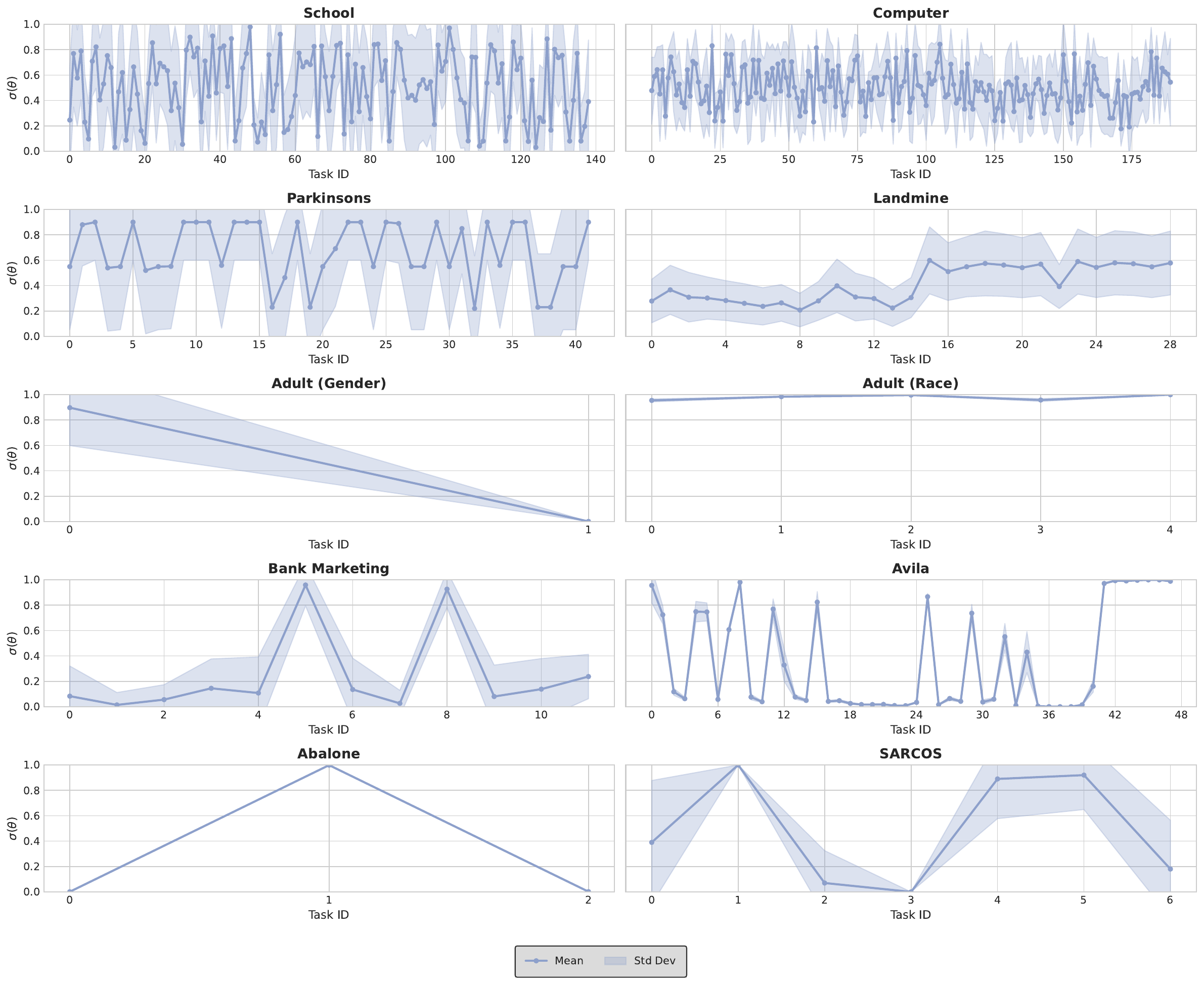}
      \caption{Mean and standard deviation of the learned $\sigma(\boldsymbol{\theta})$
            values across multiple
            runs for each task, shown separately for each dataset in the corresponding subplots.
            Values of $\sigma(\boldsymbol{\theta})$ close to $0$ or $1$ indicate a clear separation
            between non-outlier and outlier tasks, with the specific
            direction depending on initialization and alignment.
            The shaded areas represent the variability across runs. These
            results demonstrate that R-MTGB consistently identifies and
            distinguishes outlier tasks from non-outlier tasks, even
            in heterogeneous and noisy datasets.}

      \label{fig:sigmoid_thetas_real}
\end{figure}

To evaluate the effectiveness of the proposed model
in identifying outlier tasks, we analyzed the
average optimized \(\sigma({\theta}_t)\)
value for each task $t$ across the experimental datasets.
These \(\sigma(\boldsymbol{\theta})\) values serve as task-specific
outlier weights. After optimization, a value close to one indicates
that the corresponding task is likely an outlier, whereas a value
near zero suggests the task is likely a non-outlier or vice versa.
Figure~\ref{fig:sigmoid_thetas_real} shows the average
learned \(\boldsymbol{\theta}\) vector parameter by
\ac{R-MTGB} model, across $100$ runs for each dataset (subplots)
alongside the standard deviation (shaded region) for each task.
To ensure consistent directionality across different experiment
runs, each vector
\(\sigma(\boldsymbol{\theta})\) is aligned with a reference
vector (taken from the first run).
Specifically, the first vector is stored as the reference.
For each subsequent vector,
the correlation with the reference is checked.
If the correlation is negative
(indicating opposite directionality) the vector is flipped by taking
\(1 - \sigma(\boldsymbol{\theta})\). This operation ensures
that all vectors point in
the same direction in the latent space,
eliminating ambiguity due to symmetry.
For datasets containing distinguishable or noisy tasks,
such as \textit{Avila}, \textit{School},
and \textit{Bank Marketing},
\ac{R-MTGB} consistently assigned \(\sigma(\boldsymbol{\theta})\) values near the
extremes, reflecting confident separation between non-outlier and
outlier tasks. In contrast, \textit{Adult}s datasets exhibit
minimal variation in \(\sigma(\boldsymbol{\theta})\), which resulted
in more uniform distributed
\(\sigma(\boldsymbol{\theta})\) values.
Moreover, the small standard deviation across tasks for complex
datasets (e.g., \textit{SARCOS}, \textit{Avila}, and \textit{Abalone}),
indicates the robustness of the proposed model
in identifying outlier tasks in various runs.

An additional experiment examining
the training time of the studied models is
presented in~\ref{appendix_b}. The empirical results
show that, although \ac{R-MTGB} incurs a moderately
higher computational cost than \ac{ST}-\ac{GB} and
pooling-based approaches due to its joint optimization
and outlier detection mechanism, its
training time remains comparable to that of
standard \ac{MTGB}.
Across datasets, \ac{R-MTGB}
scales efficiently with increasing task numbers and
dataset size, demonstrating stable and practical
runtime performance despite its enhanced robustness.

\section{Conclusions}
\label{sec:conclusions}
This study introduced \acf{R-MTGB}, a principled methodology
comprising three blocks designed to address task
heterogeneity in \acf{MTL}.
\ac{R-MTGB} sequentially integrates shared-level
knowledge, outlier-aware task partitioning,
and task-specific fine-tuning, to build a composite
prediction model that effectively balances generalization
and specialization. Its ensemble-based formulation employs
a learned, task-dependent weighting
mechanism to adaptively interpolate between outlier and
non-outlier components, ensuring robust performance
even in the presence of anomalous tasks.
Notably, the model offers interpretability by revealing task-level
outlier scores via the learned interpolation parameters,
allowing for the diagnosis and visualization of tasks that
significantly deviate from the shared structure.

Comprehensive experiments conducted on both
synthetic and real-world datasets demonstrate the ability of the
proposed model to generalize across tasks,
maintain high predictive accuracy for each individual task,
and robustly identify anomalous tasks.
The results indicate that \ac{R-MTGB} outperforms
\acf{MTGB}, \acf{ST} learning, and \acf{DP} approaches,
including augmented data pooling with task-specific information.
Our experiments show that the
proposed method offers advantages in both settings,
though they are more pronounced in the regression setting than
in classification problems.
These advantages hold across varying degrees of task heterogeneity,
underscoring the robustness and adaptability of the developed
\ac{MTL} framework.

Finally, the proposed three-block structure,
along with the learnable parameter, is both
empirically validated and theoretically analyzed and bounded.

Future work could extend \ac{R-MTGB} to jointly learn all blocks,
rather than sequentially, which may improve optimality by
allowing the model to reconsider the inclusion or exclusion of
base learners across blocks. Currently, once base learners
are incorporated in the first block, they cannot be removed even
if they prove unnecessary in later blocks. Additionally,
the method could be extended to handle multiple task groups
beyond the current outlier/non-outlier distinction,
enabling it to automatically identify clusters of
related tasks while isolating unrelated ones.

\section*{Declaration of Interests}
The authors declare that there are no competing financial
interests or personal relationships that could have potentially
biased the research, experiments, or the conclusions
presented in this manuscript.

\section*{Acknowledgments}
The authors acknowledge financial support
from the project
PID2022-139856NB-I00,
funded by MCIN\slash AEI\slash 10.13039\slash
501100011033\slash FEDER, UE;
from project IDEA-CM (TEC-2024\slash COM-89),
funded by the Autonomous Community of Madrid;
and from the ELLIS Unit Madrid.
The authors also acknowledge computational support
from the Centro de Computaci\'on Científica-Universidad
Aut\'onoma de Madrid (CCC-UAM).

\section*{Data and Code Availability}
The datasets utilized in this study were
obtained from their respective publicly cited references.
Both these datasets and the public source developed code for
the proposed model are accessible at
\href{https://github.com/GAA-UAM/R-MTGB}
{github.com/GAA-UAM/R-MTGB}.

\appendix
\section{Additional Experiments}
\label{appendix_a}

In this appendix, we report additional experiments to
further enrich the evaluation of the studied models.
Table~\ref{tab:f1} presents the testing macro F1
scores across datasets and models.
For each run, the F1 score was computed using
the \emph{macro} averaging scheme, which calculates
the F1 score independently for each class
and then takes the unweighted mean.

Scores were first averaged over tasks within
each batch and then across repetitions, with both
the mean and standard deviation reported in the Table~\ref{tab:f1}.
As shown in Table~\ref{tab:f1}, \ac{R-MTGB}
achieves the highest or near-highest macro F1 scores on
most datasets, notably \textit{Adult (Gender)}, \textit{Adult (Race)},
and \textit{Bank Marketing}.
For the \textit{Avila} and \textit{Landmine} datasets,
\ac{ST}-\ac{GB} baseline slightly outperforms \ac{R-MTGB}.

\begin{table}[ht]
      \centering
      \footnotesize
      \caption{Testing F1 score across models for each dataset,
            averaged first over tasks within each batch
            and then over runs. Mean and standard deviation
            are reported, with best values per dataset shown in bold.}
      \setlength{\extrarowheight}{2pt}
      \begin{tabularx}{\textwidth}{l
                  >{\centering\arraybackslash}X
                  >{\centering\arraybackslash}X
                  >{\centering\arraybackslash}X
                  >{\centering\arraybackslash}X
                  >{\centering\arraybackslash}X}
            \toprule
            \textbf{Model}           & \textbf{Adult (Gender)}               & \textbf{Adult (Race)}                 & \textbf{Avila}                        & \textbf{Bank Marketing}               & \textbf{Landmine}                     \\
            \midrule
            \textbf{\textsc{R-MTGB}} & \textbf{0.7574} $\pm$ \textbf{0.0058} & \textbf{0.7565} $\pm$ \textbf{0.0054} & 0.4595 $\pm$ 0.0969                   & \textbf{0.6198} $\pm$ \textbf{0.0126} & 0.5713 $\pm$ 0.0163                   \\
            \textbf{\textsc{MTGB}}   & 0.7531 $\pm$ 0.0055                   & 0.7477 $\pm$ 0.0074                   & 0.4605 $\pm$ 0.1014                   & 0.6024 $\pm$ 0.0137                   & 0.5725 $\pm$ 0.0165                   \\
            \textbf{\textsc{DP-GB}}  & 0.7173 $\pm$ 0.0107                   & 0.7173 $\pm$ 0.0107                   & 0.1442 $\pm$ 0.0062                   & 0.5317 $\pm$ 0.0067                   & 0.4842 $\pm$ 0.0009                   \\
            \textbf{\textsc{ST-GB}}  & 0.7369 $\pm$ 0.0060                   & 0.7263 $\pm$ 0.0080                   & \textbf{0.4629} $\pm$ \textbf{0.1087} & 0.5711 $\pm$ 0.0082                   & \textbf{0.5772} $\pm$ \textbf{0.0155} \\
            \textbf{\textsc{TaF-GB}} & 0.7173 $\pm$ 0.0107                   & 0.7173 $\pm$ 0.0107                   & 0.1466 $\pm$ 0.0058                   & 0.5317 $\pm$ 0.0067                   & 0.4842 $\pm$ 0.0009                   \\
            \bottomrule
      \end{tabularx}
      \label{tab:f1}
\end{table}

To extend the experiment to a deep learning framework,
we incorporated a \ac{DNN} model into an additional
batch of experiments.
The \ac{DNN} was trained on the pooled data
from all tasks, allowing the model to learn a
shared representation that captures
general patterns across.
This configuration serves as a deep-learning
benchmark that leverages cross-task
information to enhance representation learning and
allows a direct empirical comparison with
the boosting-based \ac{MTL} models.

The trained \ac{DNN} architecture
consists of three fully connected hidden layers with $100$
neurons each, using the Rectified Linear Unit (ReLU) as
the activation function. Training was performed for a
maximum of 100 epochs with an L2 regularization parameter of
0.0001. A grid search over the learning rate initialization
values ${[0.001, 0.01, 0.1]}$ was conducted for hyperparameter
tuning using 5-fold cross-validation. All input features
were standardized through a preprocessing pipeline, and model
training was conducted using the \verb?scikit-learn? library.

The \ac{DNN} was trained on the same $100$ distinct synthetic
train/test datasets described in
Subsection~\ref{sub_sec:synthetic_experiments}.
summarized in Tables~\ref{tab:recall_accuracy_appendix} and
\ref{tab:mae_rmse_appendix}, provides a complementary
benchmark to the boosting-based \ac{MTL} models presented earlier
(See Subsection~\ref{sub_sec:synthetic_experiments}).

As shown in Table~\ref{tab:recall_accuracy_appendix},
\ac{DNN} attains competitive classification performance.
Its test accuracy and recall are close to those of \ac{R-MTGB},
(see Table~\ref{tab:recall_accuracy}), placing \ac{R-MTGB}
as the top-performing model in terms of accuracy and
the second-best in recall, with an insignificant difference
between the two.
\begin{table}[ht]
      \centering
      \footnotesize
      \caption{Average recall and accuracy scores of \ac{DNN} with standard
            deviations, computed by first averaging across tasks
            and then over runs.}
      \begin{tabularx}{\textwidth}{lXXXX}
            \toprule
            \textbf{Model}        & \multicolumn{2}{c}{\textbf{Recall}} & \multicolumn{2}{c}{\textbf{Accuracy}}                                           \\
                                  & \textbf{Train}                      & \textbf{Test}                         & \textbf{Train}      & \textbf{Test}     \\
            \midrule
            \textbf{\textsc{DNN}} & 0.871 $\pm$ 0.117                   & {0.830 $\pm$ 0.118}                   & {0.880 $\pm$ 0.043} & 0.841 $\pm$ 0.034 \\
            \bottomrule
      \end{tabularx}
      \label{tab:recall_accuracy_appendix}
\end{table}

According to Table~\ref{tab:mae_rmse_appendix}
and Table~\ref{tab:mae_rmse}, \ac{R-MTGB} achieves the best
generalization performance with the lowest test \ac{MAE} and \ac{RMSE}.
The performance gap between \ac{DNN} and the proposed model
highlights the advantage of \ac{R-MTGB} in capturing inter-task
structure and mitigating the impact of outlier tasks,
confirming its superior robustness in heterogeneous
multi-task settings.

\begin{table}[ht]
      \centering
      \footnotesize
      \caption{Average \ac{MAE} and \ac{RMSE} scores of \ac{DNN} with standard
            deviations, computed by first averaging across tasks
            and then over runs.}
      \begin{tabularx}{\textwidth}{lXXXX}
            \toprule
            \textbf{Model}        & \multicolumn{2}{c}{\textbf{MAE}} & \multicolumn{2}{c}{\textbf{RMSE}}                                         \\
                                  & \textbf{Train}                   & \textbf{Test}                     & \textbf{Train}    & \textbf{Test}     \\
            \midrule
            \textbf{\textsc{DNN}} & {0.323} $\pm$ 0.113              & 0.380 $\pm$ 0.105                 & 0.458 $\pm$ 0.145 & 0.546 $\pm$ 0.141 \\
            \bottomrule
      \end{tabularx}
      \label{tab:mae_rmse_appendix}
\end{table}

\begin{figure}[!htbp]
      \centering
      \includegraphics[width=\linewidth]{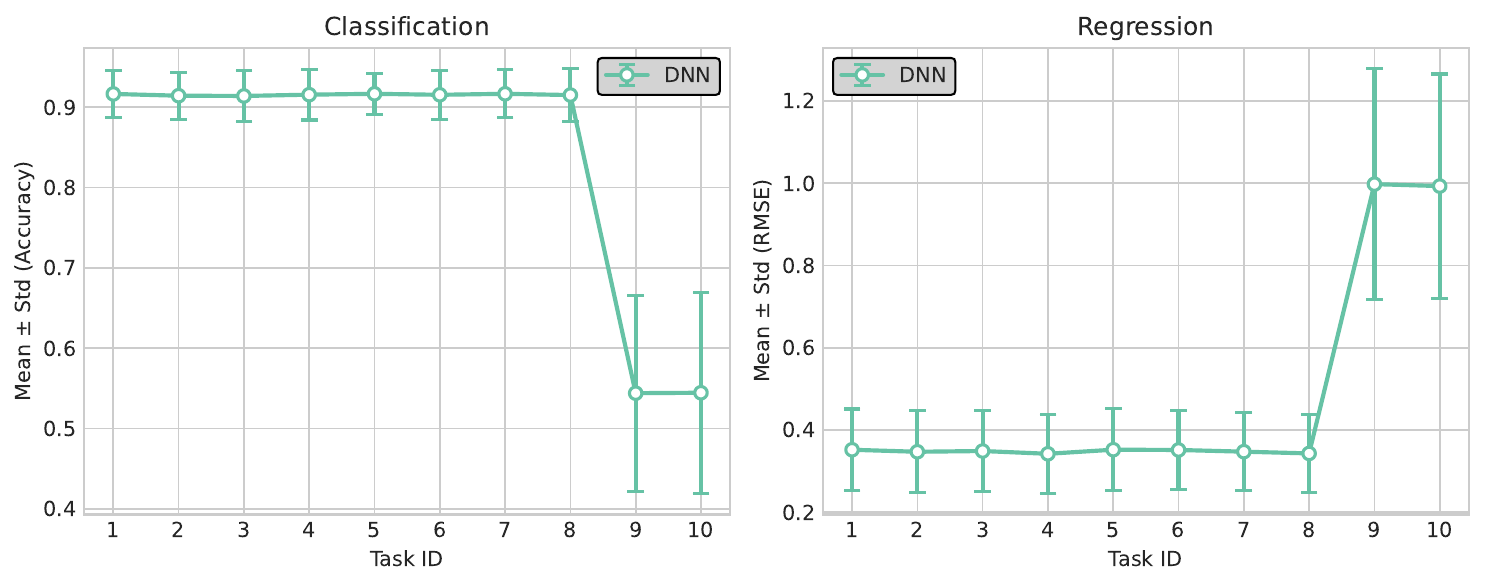}
      \caption{Average task-wise performance of the
            \ac{DNN} over multiple runs shown
            separately for classification (left subplot)
            and regression (right subplot) tasks.}
      \label{fig:dnn_per_task}
\end{figure}

To further examine the performance of \ac{DNN} across tasks,
Figure~\ref{fig:dnn_per_task}
presents the average \ac{DNN} results per task across all
repetitions—analogous to Figure~\ref{fig:performance_across_tasks_synthetic_data}.
Specifically, the left subplot of Figure~\ref{fig:dnn_per_task}
illustrates classification accuracy,
while the right subplot displays regression performance
measured by \ac{RMSE}. Comparing Figures~\ref{fig:dnn_per_task}
and~\ref{fig:performance_across_tasks_synthetic_data} (left subplots),
we observe that \ac{DNN} achieves better results on non-outlier
tasks (tasks 1 to 8) relative to the proposed models,
but struggles with outlier tasks. This is
particularly evident in the last two tasks (Tasks~9 and~10),
where accuracy drops significantly.

For regression (right subplots of Figure~\ref{fig:dnn_per_task}
and Figure~\ref{fig:performance_across_tasks_synthetic_data}),
\ac{DNN} achieves comparable performance to the
proposed model on non-outlier tasks, but again
fails to generalize to outlier tasks, resulting in large
errors and a substantial performance gap relative to
the proposed approach.

Given the performance of \ac{DNN} on the synthetic datasets,
we omit experiments using \ac{DNN}
on real-world datasets, as the complex synthetic setup
already provides a sufficient basis for comparison.

\section{Complexity and Training Efficiency}
\label{appendix_b}

In this section, we provide both theoretical and empirical
analyses of the computational complexity of the models
studied in this paper. For the theoretical analysis,
we employ Big~$\mathcal{O}$
notation to describe the time complexity of each model.

Let \(N = \sum_{t=1}^{T} N_t\) be the pooled number of samples,
\(d\) the number of features,
and \(\text{Tree}(n,d)\) the cost of fitting one decision stump
on \(n\) samples with \(d\) features.
For stumps we approximate
\(\text{Tree}(n,d) = \mathcal{O}(n d \log n)\).
We additionally assume that all tasks have similar size,
i.e. \(N_t \approx N/T\).

\begin{itemize}

      \item \textbf{\ac{ST}-\ac{GB}:}
            trains one model per task with \(M_3\) trees each:
            \begin{equation*}
                  \begin{aligned}
                        \mathcal{O}\Big(
                        \sum_{t=1}^{T} M_3\,\text{Tree}(N_t,d)
                        \Big)
                         & \approx
                        \mathcal{O}\Big(
                        T M_3\,\text{Tree}(N/T,d)
                        \Big)      \\
                         & =
                        \mathcal{O}\Big(
                        M_3\, N d \log(N/T)
                        \Big).
                  \end{aligned}
            \end{equation*}

      \item \textbf{\ac{MTGB} (shared + per-task):}
            Block~1 fits one multi-output stump per iteration on pooled data
            (not one per task).
            \begin{equation*}
                  \begin{aligned}
                        \mathcal{O}\Big(
                        M_1\,\text{Tree}(N,d)
                        + T M_3\,\text{Tree}(N/T,d)
                        \Big)
                         & \\
                        \approx \mathcal{O}\Big(
                        M_1\, N d \log N
                        +\, M_3\, N d \log(N/T)
                        \Big).
                  \end{aligned}
            \end{equation*}

      \item \textbf{\ac{R-MTGB}:}
            Block~2 adds two pooled trees per iteration.
            Updating \(\theta_t\) costs only
            \(c_\theta(N_t)=\mathcal{O}(N_t)\),
            negligible relative to \(\text{Tree}(N,d)\).
            \begin{equation*}
                  \begin{aligned}
                        \text{Block 2:}\quad
                        \mathcal{O}\big(
                        2 M_2\,\text{Tree}(N,d)
                        \big)
                         & \\
                        = \mathcal{O}\big(
                        2 M_2\, N d \log N
                        \big).
                  \end{aligned}
            \end{equation*}
            Combining the three blocks yields
            \begin{equation*}
                  \begin{aligned}
                        \mathcal{O}\Big( M_1\,\text{Tree}(N,d)
                        + 2 M_2\,\text{Tree}(N,d)
                        + T M_3\,\text{Tree}(N/T,d) \Big)
                         & \\
                        \approx \mathcal{O}\Big( (M_1 + 2M_2)\, N d \log N
                        +\, M_3\, N d \log(N/T)
                        \Big).
                  \end{aligned}
            \end{equation*}
\end{itemize}

Under the equal-task-size assumption and stump model,
all methods exhibit the same \emph{highest-order complexity},
differing only in constant factors and in how the
iterations \(M_1, M_2, M_3\) are distributed across blocks.

For the empirical analysis, all models were
trained using their default hyperparameter settings,
as described in Section~\ref{sec:experiments}.
The number of estimators per block and per model was
tuned using the grid specified in Table~\ref{table:hparams}.
Two representative datasets (\textit{Adult (Gender)}) for
classification and \textit{Parkinsons}
for regression) were selected for the experiments.
Each training procedure was repeated independently five times,
and the mean and standard deviation of the results are
reported in Table~\ref{tab:training_time}, with the lowest
elapsed time highlighted in \textbf{bold}.
Elapsed times were measured on a Linux-based system using
CPU processing time. The experiments were conducted on
a machine equipped with two Intel(R) Xeon(R) E5-2620 v3 CPUs
(2.40 GHz, 6 cores per socket, 24 threads in total)
and 64 GB of RAM. The reported times correspond to
the sum of system CPU time consumed by each training
process, excluding any sleep or idle time.
The elapsed training time was recorded in
seconds by comparing timestamps before
and after the training phase.

\begin{table}[htbp]
      \centering
      \scriptsize
      \setlength{\tabcolsep}{3pt}
      \renewcommand{\arraystretch}{1.2}
      \caption{Average training time for each method,
            in seconds, including estimating the hyperparameters
            using the inner CV procedure and fitting the final model.
            Lowest times are highlighted in bold.}
      \begin{tabularx}{\textwidth}{lYY}
            \hline
            \textbf{Model}           & \textbf{Adult (Gender)}                     & \textbf{Parkinson}                          \\
            \hline
            \textbf{\textsc{R-MTGB}} & 12.9838014 $\pm$ 0.8103452                  & 4.0310959 $\pm$ 0.1997661                   \\
            \textbf{\textsc{MTGB}}   & 8.1746369 $\pm$ 0.5158048                   & 3.2083791 $\pm$ 0.1308600                   \\
            \textbf{\textsc{DP-GB}}  & 6.1319405 $\pm$ 0.3305553                   & \textbf{0.9071499} $\pm$ \textbf{0.0655788} \\
            \textbf{\textsc{ST-GB}}  & 6.4622585 $\pm$ 0.4245966                   & 2.9695122 $\pm$ 0.1109322                   \\
            \textbf{\textsc{TaF-GB}} & \textbf{5.7476897} $\pm$ \textbf{0.2548860} & 1.0148046 $\pm$ 0.0911100                   \\
            \hline
      \end{tabularx}
      \label{tab:training_time}
\end{table}

Based on the measured training times reported in
Table~\ref{tab:training_time},
models that rely on pooled data are the fastest to train.
\ac{ST}-\ac{GB} follows closely, remaining computationally
cheaper than the multi-task approaches. In contrast, \ac{MTGB}
requires more time.
Among all methods, \ac{R-MTGB} exhibits the highest training time.
Its three-block architecture, together with the
larger number of hyperparameters that must be tuned,
makes it more computationally demanding than the alternatives.
A similar, though smaller, effect is observed for \ac{MTGB},
whose additional hyperparameter increases its training cost
relative to pooling-based and single-task models.
Despite this, the difference in runtime between \ac{R-MTGB}
and \ac{MTGB} remains moderate, indicating that the
robustness gains provided by \ac{R-MTGB} are achieved
with only a reasonable computational overhead.

\bibliographystyle{plainnat}
\bibliography{references}

@misc{Becker1996,
  author       = {Becker, Barry and Kohavi, Ronny},
  title        = {{Adult}},
  year         = {1996},
  howpublished = {\url{https://archive.ics.uci.edu}}
}

@misc{Tsanas2009,
  author       = {Tsanas, Athanasios and Little, Max},
  title        = {{Parkinsons Telemonitoring}},
  year         = {2009},
  howpublished = {\url{https://archive.ics.uci.edu}}
}

@misc{Nash1994,
  author       = {Warwick Nash and Tracy Sellers and Simon Talbot and Andrew Cawthorn and Wes Ford},
  title        = {{Abalone}},
  year         = {1994},
  howpublished = {\url{https://archive.ics.uci.edu}}
}

@inproceedings{Argyriou2007,
  author    = {Argyriou, Andreas and Pontil, Massimiliano and Ying, Yiming and Micchelli, Charles},
  booktitle = {Advances in Neural Information Processing Systems},
  publisher = {Curran Associates, Inc.},
  title     = {{A Spectral Regularization Framework for Multi-Task Structure Learning}},
  volume    = {20},
  year      = {2007}
}

@article{bakker2003,
  title   = {{Task clustering and gating for bayesian multitask learning}},
  author  = {Bakker, Bart and Heskes, Tom},
  journal = {Journal of Machine Learning Research},
  volume  = {4},
  pages   = {83--99},
  year    = {2003}
}

@inproceedings{Ciliberto2017,
  author    = {Ciliberto, Carlo and Rudi, Alessandro and Rosasco, Lorenzo and Pontil, Massimiliano},
  booktitle = {Advances in Neural Information Processing Systems},
  publisher = {Curran Associates, Inc.},
  title     = {{Consistent Multitask Learning with Nonlinear Output Relations}},
  volume    = {30},
  year      = {2017}
}

@inproceedings{Emami2023,
  author    = {Emami, Seyedsaman
               and Ruiz Pastor, Carlos
               and Mart{\'i}nez-Mu{\~{n}}oz, Gonzalo},
  title     = {{Multi-Task Gradient Boosting}},
  booktitle = {Hybrid Artificial Intelligent Systems},
  year      = {2023},
  publisher = {Springer International Publishing},
  pages     = {97--107}
}

@inproceedings{Jawanpuria2015,
  author    = {Jawanpuria, Pratik Kumar and Lapin, Maksim and Hein, Matthias and Schiele, Bernt},
  booktitle = {Advances in Neural Information Processing Systems},
  publisher = {Curran Associates, Inc.},
  title     = {{Efficient Output Kernel Learning for Multiple Tasks}},
  volume    = {28},
  year      = {2015}
}

@article{Lenk1996,
  author  = {Lenk, Peter J. and DeSarbo, Wayne S. and Green, Paul E. and Young, Martin R.},
  title   = {{Hierarchical Bayes Conjoint Analysis: Recovery of Partworth Heterogeneity from Reduced Experimental Designs}},
  journal = {Marketing Science},
  volume  = {15},
  number  = {2},
  pages   = {173-191},
  year    = {1996}
}

@inproceedings{Oneto2019,
  author    = {Oneto, Luca and Doninini, Michele and Elders, Amon and Pontil, Massimiliano},
  title     = {{Taking Advantage of Multitask Learning for Fair Classification}},
  year      = {2019},
  publisher = {Association for Computing Machinery},
  booktitle = {Proceedings of the 2019 AAAI/ACM Conference on AI, Ethics, and Society},
  pages     = {227-237}
}

@inproceedings{Wang2021,
  author    = {Wang, Yuyan and Wang, Xuezhi and Beutel, Alex and Prost, Flavien and Chen, Jilin and Chi, Ed H.},
  title     = {{Understanding and Improving Fairness-Accuracy Trade-offs in Multi-Task Learning}},
  year      = {2021},
  publisher = {Association for Computing Machinery},
  booktitle = {Proceedings of the 27th ACM SIGKDD Conference on Knowledge Discovery \& Data Mining},
  pages     = {1748-1757}
}

@inproceedings{Wang2022,
  author    = {Wang, Sinan and Li, Yumeng and Li, Hongyan and Zhu, Tanchao and Li, Zhao and Ou, Wenwu},
  booktitle = {2022 IEEE 38th International Conference on Data Engineering (ICDE)},
  title     = {{Multi-Task Learning with Calibrated Mixture of Insightful Experts}},
  year      = {2022},
  pages     = {3307-3319}
}

@article{Yilmaz2018,
  author  = {Yilmaz, Cemal and Kahraman, Hamdi Tolga and Söyler, Salih},
  journal = {IEEE Access},
  title   = {{Passive Mine Detection and Classification Method Based on Hybrid Model}},
  year    = {2018},
  volume  = {6},
  pages   = {47870-47888}
}

@inproceedings{Zhao2018,
  title   = {{Data Poisoning Attacks on Multi-Task Relationship Learning}},
  volume  = {32},
  journal = {Proceedings of the AAAI Conference on Artificial Intelligence},
  author  = {Zhao, Mengchen and An, Bo and Yu, Yaodong and Liu, Sulin and Pan, Sinno},
  year    = {2018}
}

@article{Argyriou2008,
  author  = {Andreas Argyriou and Theodoros Evgeniou and Massimiliano Pontil},
  title   = {{Convex multi-task feature learning}},
  journal = {Machine Learning},
  year    = {2008},
  volume  = {73},
  number  = {3},
  pages   = {243--272}
}

@article{Srinivasan2024,
  author  = {Srinivasan, Saravanan and Ramadass, Parthasarathy and Mathivanan, Sandeep Kumar and Panneer Selvam, Karthikeyan and Shivahare, Basu Dev and Shah, Mohd Asif},
  title   = {{Detection of Parkinson disease using multiclass machine learning approach}},
  journal = {Scientific Reports},
  year    = {2024},
  volume  = {14},
  number  = {1},
  pages   = {13813}
}

@article{Gunduz2019,
  author  = {Gunduz, Hakan},
  journal = {IEEE Access},
  title   = {{Deep Learning-Based Parkinson's Disease Classification Using Vocal Feature Sets}},
  year    = {2019},
  volume  = {7},
  pages   = {115540-115551}
}

@article{Stefano2018,
  title   = {{Reliable writer identification in medieval manuscripts through page layout features: The “Avila” Bible case}},
  journal = {Engineering Applications of Artificial Intelligence},
  volume  = {72},
  pages   = {99-110},
  year    = {2018},
  author  = {C. De Stefano and M. Maniaci and F. Fontanella and A. Scotto di Freca}
}

@inproceedings{Moro2011,
  author    = {S. Moro and R. Laureano and P. Cortez},
  title     = {{Using Data Mining for Bank Direct Marketing: An Application of the CRISP-DM Methodology}},
  booktitle = {Proceedings of the European Simulation and Modelling Conference - ESM'2011},
  pages     = {117--121},
  year      = {2011},
  publisher = {EUROSIS}
}

@article{Friedman2001,
  author    = {Jerome H. Friedman},
  journal   = {The Annals of Statistics},
  number    = {5},
  pages     = {1189--1232},
  publisher = {Institute of Mathematical Statistics},
  title     = {{Greedy Function Approximation: A Gradient Boosting Machine}},
  volume    = {29},
  year      = {2001}
}

@article{demvsar2006statistical,
  title     = {{Statistical comparisons of classifiers over multiple data sets}},
  author    = {Dem{\v{s}}ar, Janez},
  journal   = {The Journal of Machine Learning Research},
  volume    = {7},
  pages     = {1--30},
  year      = {2006},
  publisher = {JMLR. org}
}

@inproceedings{Rahimi2007,
  author    = {Rahimi, Ali and Recht, Benjamin},
  booktitle = {Advances in Neural Information Processing Systems},
  publisher = {Curran Associates, Inc.},
  title     = {{Random Features for Large-Scale Kernel Machines}},
  volume    = {20},
  year      = {2007}
}

@article{pedregosa2011scikit,
  title     = {{Scikit-learn: Machine learning in Python}},
  author    = {Pedregosa, Fabian and Varoquaux, Ga{\"e}l and Gramfort, Alexandre and Michel, Vincent and Thirion, Bertrand and Grisel, Olivier and Blondel, Mathieu and Prettenhofer, Peter and Weiss, Ron and Dubourg, Vincent and others},
  journal   = {the Journal of machine Learning research},
  volume    = {12},
  pages     = {2825--2830},
  year      = {2011},
  publisher = {JMLR. org}
}

@article{Caruana1997,
  author  = {Rich Caruana},
  title   = {{Multitask Learning}},
  journal = {Machine Learning},
  year    = {1997},
  volume  = {28},
  number  = {1},
  pages   = {41--75}
}

@article{Zhang2022,
  author  = {Zhang, Yu and Yang, Qiang},
  journal = {IEEE Transactions on Knowledge and Data Engineering},
  title   = {{A Survey on Multi-Task Learning}},
  year    = {2022},
  volume  = {34},
  number  = {12},
  pages   = {5586-5609}
}

@inproceedings{Liao2005,
  author    = {Liao, Xuejun and Carin, Lawrence},
  booktitle = {Advances in Neural Information Processing Systems},
  publisher = {MIT Press},
  title     = {{Radial Basis Function Network for Multi-task Learning}},
  volume    = {18},
  year      = {2005}
}

@inproceedings{Zhang2014,
  author    = {Zhang, Zhanpeng
               and Luo, Ping
               and Loy, Chen Change
               and Tang, Xiaoou},
  title     = {{Facial Landmark Detection by Deep Multi-task Learning}},
  booktitle = {Computer Vision -- ECCV 2014},
  year      = {2014},
  publisher = {Springer International Publishing},
  pages     = {94--108}
}

@inproceedings{Liu2015,
  author    = {Liu, Wu and Mei, Tao and Zhang, Yongdong and Che, Cherry and Luo, Jiebo},
  title     = {{Multi-Task Deep Visual-Semantic Embedding for Video Thumbnail Selection}},
  booktitle = {Proceedings of the IEEE Conference on Computer Vision and Pattern Recognition (CVPR)},
  pages     = {3707--3715},
  year      = {2015}
}

@inproceedings{Zhang2015,
  author    = {Zhang, Wenlu and Li, Rongjian and Zeng, Tao and Sun, Qian and Kumar, Sudhir and Ye, Jieping and Ji, Shuiwang},
  title     = {{Deep Model Based Transfer and Multi-Task Learning for Biological Image Analysis}},
  year      = {2015},
  publisher = {Association for Computing Machinery},
  booktitle = {Proceedings of the 21th ACM SIGKDD International Conference on Knowledge Discovery and Data Mining},
  pages     = {1475-1484}
}

@inproceedings{LI2014,
  author    = {LI, Sijin and Liu, Zhi-Qiang and Chan, Antoni B.},
  title     = {{Heterogeneous Multi-task Learning for Human Pose Estimation with Deep Convolutional Neural Network}},
  booktitle = {Proceedings of the IEEE Conference on Computer Vision and Pattern Recognition (CVPR) Workshops},
  pages     = {482--489},
  year      = {2014}
}

@article{ando2005framework,
  title   = {{A framework for learning predictive structures from multiple tasks and unlabeled data}},
  author  = {Ando, Rie Kubota and Zhang, Tong and Bartlett, Peter},
  journal = {Journal of machine learning research},
  volume  = {6},
  number  = {11},
  year    = {2005}
}

@inproceedings{Agiza2024,
  author    = {Agiza, Ahmed and Neseem, Marina and Reda, Sherief},
  title     = {{MTLoRA: Low-Rank Adaptation Approach for Efficient Multi-Task Learning}},
  booktitle = {Proceedings of the IEEE/CVF Conference on Computer Vision and Pattern Recognition (CVPR)},
  year      = {2024},
  pages     = {16196-16205}
}

@inproceedings{Chen2011,
  author    = {Chen, Jianhui and Zhou, Jiayu and Ye, Jieping},
  title     = {{Integrating low-rank and group-sparse structures for robust multi-task learning}},
  year      = {2011},
  publisher = {Association for Computing Machinery},
  booktitle = {Proceedings of the 17th ACM SIGKDD International Conference on Knowledge Discovery and Data Mining},
  pages     = {42-50}
}

@inproceedings{Han2016,
  title     = {{Multi-Stage Multi-Task Learning with Reduced Rank}},
  volume    = {30},
  booktitle = {Proceedings of the AAAI Conference on Artificial Intelligence},
  author    = {Han, Lei and Zhang, Yu},
  year      = {2016}
}

@inproceedings{Thrun1996,
  title        = {{Discovering structure in multiple learning tasks: The TC algorithm}},
  author       = {Thrun, Sebastian and O'Sullivan, Joseph},
  booktitle    = {ICML},
  volume       = {96},
  pages        = {489--497},
  year         = {1996},
  organization = {Citeseer}
}

@inproceedings{Crammer2012,
  author    = {Crammer, Koby and Mansour, Yishay},
  booktitle = {Advances in Neural Information Processing Systems},
  publisher = {Curran Associates, Inc.},
  title     = {{Learning Multiple Tasks using Shared Hypotheses}},
  volume    = {25},
  year      = {2012}
}

@inproceedings{Evgeniou2004,
  author    = {Evgeniou, Theodoros and Pontil, Massimiliano},
  title     = {{Regularized multi--task learning}},
  year      = {2004},
  publisher = {Association for Computing Machinery},
  booktitle = {Proceedings of the Tenth ACM SIGKDD International Conference on Knowledge Discovery and Data Mining},
  pages     = {109-117}
}

@inproceedings{Parameswaran2010,
  author    = {Parameswaran, Shibin and Weinberger, Kilian Q},
  booktitle = {Advances in Neural Information Processing Systems},
  publisher = {Curran Associates, Inc.},
  title     = {{Large Margin Multi-Task Metric Learning}},
  volume    = {23},
  year      = {2010}
}

@article{Evgeniou2005,
  title   = {{Learning multiple tasks with kernel methods}},
  author  = {Evgeniou, Theodoros and Micchelli, Charles A and Pontil, Massimiliano and Shawe-Taylor, John},
  journal = {Journal of machine learning research},
  volume  = {6},
  number  = {4},
  year    = {2005}
}

@inproceedings{Kato2007,
  author    = {Kato, Tsuyoshi and Kashima, Hisashi and Sugiyama, Masashi and Asai, Kiyoshi},
  booktitle = {Advances in Neural Information Processing Systems},
  publisher = {Curran Associates, Inc.},
  title     = {{Multi-Task Learning via Conic Programming}},
  volume    = {20},
  year      = {2007}
}

@inproceedings{Han2015,
  author    = {Han, Lei and Zhang, Yu},
  title     = {{Learning Tree Structure in Multi-Task Learning}},
  year      = {2015},
  publisher = {Association for Computing Machinery},
  booktitle = {Proceedings of the 21th ACM SIGKDD International Conference on Knowledge Discovery and Data Mining},
  pages     = {397-406}
}

@inproceedings{Shen2024,
  author    = {Shen, Sheng and Yang, Shijia and Zhang, Tianjun and Zhai, Bohan and Gonzalez, Joseph E. and Keutzer, Kurt and Darrell, Trevor},
  title     = {{Multitask Vision-Language Prompt Tuning}},
  booktitle = {Proceedings of the IEEE/CVF Winter Conference on Applications of Computer Vision (WACV)},
  year      = {2024},
  pages     = {5656-5667}
}

@inproceedings{Liu2024,
  author    = {Liu, Qidong and Wu, Xian and Zhao, Xiangyu and Zhu, Yuanshao and Xu, Derong and Tian, Feng and Zheng, Yefeng},
  title     = {{When MOE Meets LLMs: Parameter Efficient Fine-tuning for Multi-task Medical Applications}},
  year      = {2024},
  publisher = {Association for Computing Machinery},
  booktitle = {Proceedings of the 47th International ACM SIGIR Conference on Research and Development in Information Retrieval},
  pages     = {1104-1114}
}

@article{Tsai2025,
  author  = {Hsinhan Tsai and Ta-Wei Yang and Tien-Yi Wu and Ya-Chi Tu and Cheng-Lung Chen and Cheng-Fu Chou},
  title   = {{Multitask learning multimodal network for chronic disease prediction}},
  journal = {Scientific Reports},
  volume  = {15},
  number  = {1},
  pages   = {15468},
  year    = {2025}
}

@article{Souček2024,
  author  = {Souček, Tomáš and Alayrac, Jean-Baptiste and Miech, Antoine and Laptev, Ivan and Sivic, Josef},
  journal = {IEEE Transactions on Pattern Analysis and Machine Intelligence},
  title   = {{Multi-Task Learning of Object States and State-Modifying Actions From Web Videos}},
  year    = {2024},
  volume  = {46},
  number  = {7},
  pages   = {5114-5130}
}

@article{Bentejac2021,
  author  = {Candice Bentéjac and Anna Csörgő and Gonzalo Martínez-Muñoz},
  title   = {{A comparative analysis of gradient boosting algorithms}},
  journal = {Artificial Intelligence Review},
  volume  = {54},
  number  = {3},
  pages   = {1937--1967},
  year    = {2021}
}

@inproceedings{Chen2016,
  author    = {Chen, Tianqi and Guestrin, Carlos},
  title     = {{XGBoost: A Scalable Tree Boosting System}},
  year      = {2016},
  publisher = {Association for Computing Machinery},
  booktitle = {Proceedings of the 22nd ACM SIGKDD International Conference on Knowledge Discovery and Data Mining},
  pages     = {785-794}
}

@inproceedings{Ke2017,
  author    = {Ke, Guolin and Meng, Qi and Finley, Thomas and Wang, Taifeng and Chen, Wei and Ma, Weidong and Ye, Qiwei and Liu, Tie-Yan},
  booktitle = {Advances in Neural Information Processing Systems},
  publisher = {Curran Associates, Inc.},
  title     = {{LightGBM: A Highly Efficient Gradient Boosting Decision Tree}},
  volume    = {30},
  year      = {2017}
}

@inproceedings{Prokhorenkova2018,
  author    = {Prokhorenkova, Liudmila and Gusev, Gleb and Vorobev, Aleksandr and Dorogush, Anna Veronika and Gulin, Andrey},
  booktitle = {Advances in Neural Information Processing Systems},
  publisher = {Curran Associates, Inc.},
  title     = {{CatBoost: unbiased boosting with categorical features}},
  volume    = {31},
  year      = {2018}
}

@article{Zhang2021,
  author  = {Zhang, Zhendong and Jung, Cheolkon},
  journal = {IEEE Transactions on Neural Networks and Learning Systems},
  title   = {{GBDT-MO: Gradient-Boosted Decision Trees for Multiple Outputs}},
  year    = {2021},
  volume  = {32},
  number  = {7},
  pages   = {3156-3167}
}

@article{Emami2025,
  author  = {Seyedsaman Emami and Gonzalo Mart{\'i}nez-Mu{\~n}oz},
  title   = {{Condensed-gradient boosting}},
  journal = {International Journal of Machine Learning and Cybernetics},
  year    = {2025},
  volume  = {16},
  number  = {1},
  pages   = {687--701}
}

@article{Olivier2011,
  author  = {Olivier Chapelle and Pannagadatta Shivaswamy and Srinivas Vadrevu and Kilian Weinberger and Ya Zhang and Belle Tseng},
  title   = {{Boosted multi-task learning}},
  journal = {Machine Learning},
  year    = {2011},
  volume  = {85},
  number  = {1-2},
  pages   = {149--173}
}

@article{Jiang2020,
  title     = {{Boosting tree-assisted multitask deep learning for small scientific datasets}},
  author    = {Jiang, Jian and Wang, Rui and Wang, Menglun and Gao, Kaifu and Nguyen, Duc Duy and Wei, Guo-Wei},
  journal   = {Journal of chemical information and modeling},
  volume    = {60},
  number    = {3},
  pages     = {1235--1244},
  year      = {2020},
  publisher = {ACS Publications}
}

@article{Liu_Haizhou2024,
  author  = {Liu, Haizhou and Zhang, Xuan and Sun, Hongbin and Shahidehpour, Mohammad},
  journal = {IEEE Transactions on Smart Grid},
  title   = {{Boosted Multi-Task Learning for Inter-District Collaborative Load Forecasting}},
  year    = {2024},
  volume  = {15},
  number  = {1},
  pages   = {973-986}
}

@inproceedings{Mingcheng2021,
  author    = {Chen, Mingcheng and Wang, Zhenghui and Zhao, Zhiyun and Zhang, Weinan and Guo, Xiawei and Shen, Jian and Qu, Yanru and Lu, Jieli and Xu, Min and Xu, Yu and Wang, Tiange and Li, Mian and Tu, Weiwei and Yu, Yong and Bi, Yufang and Wang, Weiqing and Ning, Guang},
  title     = {{Task-wise Split Gradient Boosting Trees for Multi-center Diabetes Prediction}},
  year      = {2021},
  booktitle = {Proceedings of the 27th ACM SIGKDD Conference on Knowledge Discovery \& Data Mining},
  pages     = {2663-2673}
}

@article{Handong2022,
  title   = {A gradient boosting tree model for multi-department venous thromboembolism risk assessment with imbalanced data},
  journal = {Journal of Biomedical Informatics},
  volume  = {134},
  pages   = {104210},
  year    = {2022},
  author  = {Handong Ma and Zhecheng Dong and Mingcheng Chen and Wenbo Sheng and Yao Li and Weinan Zhang and Shaodian Zhang and Yong Yu}
}

@misc{ZhenZhe2022,
  title        = {{MT-GBM: A Multi-Task Gradient Boosting Machine with Shared Decision Trees}},
  author       = {ZhenZhe Ying and Zhuoer Xu and Zhifeng Li and Weiqiang Wang and Changhua Meng},
  year         = {2022},
  eprint       = {2201.06239},
  howpublished = {arXiv}
}

@article{Gunasekara2024,
  author  = {Nuwan Gunasekara and Bernhard Pfahringer and Heitor Gomes and Albert Bifet},
  title   = {{Gradient boosted trees for evolving data streams}},
  journal = {Machine Learning},
  volume  = {113},
  number  = {5},
  pages   = {3325--3352},
  year    = {2024}
}

@article{Wenchao2024,
  title   = {{A novel gradient boosting approach for imbalanced regression}},
  journal = {Neurocomputing},
  volume  = {601},
  pages   = {128091},
  year    = {2024},
  author  = {Wenchao Zhang and Peixin Shi and Pengjiao Jia and Xiaoqi Zhou}
}

@article{Plaia2022,
  author  = {Antonella Plaia and Simona Buscemi and Johannes F{\"u}rnkranz and Eneldo Loza Menc{\'i}a},
  title   = {{Comparing Boosting and Bagging for Decision Trees of Rankings}},
  journal = {Journal of Classification},
  volume  = {39},
  number  = {1},
  pages   = {78--99},
  year    = {2022}
}

@article{Manar2022,
  title   = {{Missing value estimation using clustering and deep learning within multiple imputation framework}},
  journal = {Knowledge-Based Systems},
  volume  = {249},
  pages   = {108968},
  year    = {2022},
  author  = {Manar D. Samad and Sakib Abrar and Norou Diawara}
}

@article{Samir2018,
  title   = {{Gradient boosting machine for modeling the energy consumption of commercial buildings}},
  journal = {Energy and Buildings},
  volume  = {158},
  pages   = {1533-1543},
  year    = {2018},
  author  = {Samir Touzani and Jessica Granderson and Samuel Fernandes}
}

@article{Taha2020,
  author  = {Taha, Altyeb Altaher and Malebary, Sharaf Jameel},
  journal = {IEEE Access},
  title   = {{An Intelligent Approach to Credit Card Fraud Detection Using an Optimized Light Gradient Boosting Machine}},
  year    = {2020},
  volume  = {8},
  pages   = {25579-25587}
}

@article{Natekin2013,
  author  = {Natekin, Alexey  and Knoll, Alois },
  title   = {{Gradient boosting machines, a tutorial}},
  journal = {Frontiers in Neurorobotics},
  volume  = {Volume 7 - 2013},
  year    = {2013}
}

@article{Ravid2022,
  title   = {{Tabular data: Deep learning is not all you need}},
  journal = {Information Fusion},
  volume  = {81},
  pages   = {84-90},
  year    = {2022},
  author  = {Ravid Shwartz-Ziv and Amitai Armon}
}

@inproceedings{Gong2012,
  title     = {{Robust multi-task feature learning}},
  author    = {Gong, Pinghua and Ye, Jieping and Zhang, Changshui},
  booktitle = {Proceedings of the 18th ACM SIGKDD international conference on Knowledge discovery and data mining},
  pages     = {895--903},
  year      = {2012}
}

@inproceedings{Yu2007,
  author    = {Yu, Shipeng and Tresp, Volker and Yu, Kai},
  title     = {{Robust multi-task learning with t-processes}},
  year      = {2007},
  publisher = {Association for Computing Machinery},
  booktitle = {Proceedings of the 24th International Conference on Machine Learning},
  pages     = {1103-1110}
}

@inproceedings{zhang2012mtboost,
  author    = {Yu Zhang and Dit-Yan Yeung},
  title     = {{Multi-Task Boosting by Exploiting Task Relationships}},
  booktitle = {Machine Learning and Knowledge Discovery in Databases},
  pages     = {697--710},
  year      = {2012},
  publisher = {Springer Berlin Heidelberg}
}

@inproceedings{Bellot2018,
  author    = {Bellot, Alexis and van der Schaar, Mihaela},
  booktitle = {Advances in Neural Information Processing Systems},
  publisher = {Curran Associates, Inc.},
  title     = {{Multitask Boosting for Survival Analysis with Competing Risks}},
  volume    = {31},
  year      = {2018}
}

@inproceedings{Lakshminarayanan2017,
  author    = {Lakshminarayanan, Balaji and Pritzel, Alexander and Blundell, Charles},
  booktitle = {Advances in Neural Information Processing Systems},
  publisher = {Curran Associates, Inc.},
  title     = {{Simple and Scalable Predictive Uncertainty Estimation using Deep Ensembles}},
  volume    = {30},
  year      = {2017}
}

@inproceedings{Xia_2023_ICCV,
  author    = {Xia, Guoxuan and Bouganis, Christos-Savvas},
  title     = {{Window-Based Early-Exit Cascades for Uncertainty Estimation: When Deep Ensembles are More Efficient than Single Models}},
  booktitle = {Proceedings of the IEEE/CVF International Conference on Computer Vision (ICCV)},
  year      = {2023},
  pages     = {17368-17380}
}

@book{zhou2012ensemble,
  title     = {{Ensemble methods: foundations and algorithms}},
  author    = {Zhou, Zhi-Hua},
  year      = {2012},
  publisher = {CRC press}
}

@inproceedings{li2015multi,
  title     = {{Multi-Task Model and Feature Joint Learning}},
  author    = {Li, Ya and Tian, Xinmei and Liu, Tongliang and Tao, Dacheng},
  booktitle = {IJCAI},
  pages     = {3643--3649},
  year      = {2015}
}

@inproceedings{wilson2020efficiently,
  title     = {{Efficiently sampling functions from {G}aussian process posteriors}},
  author    = {Wilson, James and Borovitskiy, Viacheslav and Terenin, Alexander and Mostowsky, Peter and Deisenroth, Marc},
  booktitle = {International Conference on Machine Learning},
  pages     = {10292--10302},
  year      = {2020}
}

@article{Aybike2024,
  title   = {{RAGN-L: A stacked ensemble learning technique for classification of Fire-Resistant columns}},
  journal = {Expert Systems with Applications},
  volume  = {240},
  pages   = {122491},
  year    = {2024},
  author  = {Aybike {Özyüksel Çiftçioğlu}}
}

@article{Maciej2016,
  title   = {{Ensemble boosted trees with synthetic features generation in application to bankruptcy prediction}},
  journal = {Expert Systems with Applications},
  volume  = {58},
  pages   = {93-101},
  year    = {2016},
  author  = {Maciej Zięba and Sebastian K. Tomczak and Jakub M. Tomczak}
}

@article{Azal2024,
  title   = {{A review of ensemble learning and data augmentation models for class imbalanced problems: Combination, implementation and evaluation}},
  journal = {Expert Systems with Applications},
  volume  = {244},
  pages   = {122778},
  year    = {2024},
  author  = {Azal Ahmad Khan and Omkar Chaudhari and Rohitash Chandra}
}

@article{Hongliang2018,
  title   = {{A novel ensemble method for credit scoring: Adaption of different imbalance ratios}},
  journal = {Expert Systems with Applications},
  volume  = {98},
  pages   = {105-117},
  year    = {2018},
  author  = {Hongliang He and Wenyu Zhang and Shuai Zhang}
}

@inbook{Cunningham2008,
  author    = {Cunningham, P{\'a}draig
               and Cord, Matthieu
               and Delany, Sarah Jane},
  title     = {{Supervised Learning}},
  booktitle = {Machine Learning Techniques for Multimedia: Case Studies on Organization and Retrieval},
  year      = {2008},
  publisher = {Springer Berlin Heidelberg},
  pages     = {21--49}
}

@article{Wang2022loss,
  author  = {Wang, Qi and Ma, Yue and Zhao, Kun and Tian, Yingjie},
  journal = {Annals of Data Science},
  number  = {2},
  pages   = {187--212},
  title   = {{A Comprehensive Survey of Loss Functions in Machine Learning}},
  volume  = {9},
  year    = {2022}
}

@inproceedings{Shinohara2016,
  title     = {{Adversarial multi-task learning of deep neural networks for robust speech recognition}},
  author    = {Shinohara, Yusuke},
  booktitle = {Interspeech},
  pages     = {2369--2372},
  year      = {2016}
}

@inproceedings{Misra2016,
  author    = {Misra, Ishan and Shrivastava, Abhinav and Gupta, Abhinav and Hebert, Martial},
  title     = {{Cross-Stitch Networks for Multi-Task Learning}},
  booktitle = {Proceedings of the IEEE Conference on Computer Vision and Pattern Recognition (CVPR)},
  month     = {June},
  year      = {2016}
}

@article{Fontana2024,
  author  = {Fontana, Maxime and Spratling, Michael and Shi, Miaojing},
  journal = {Proceedings of the IEEE},
  title   = {{When Multitask Learning Meets Partial Supervision: A Computer Vision Review}},
  year    = {2024},
  volume  = {112},
  number  = {6},
  pages   = {516-543}
}

@article{Vandenhende2022,
  author  = {Vandenhende, Simon and Georgoulis, Stamatios and Van Gansbeke, Wouter and Proesmans, Marc and Dai, Dengxin and Van Gool, Luc},
  journal = {IEEE Transactions on Pattern Analysis and Machine Intelligence},
  title   = {{Multi-Task Learning for Dense Prediction Tasks: A Survey}},
  year    = {2022},
  volume  = {44},
  number  = {7},
  pages   = {3614-3633}
}

@article{Chen2024,
  author    = {Chen, Shijie and Zhang, Yu and Yang, Qiang},
  title     = {{Multi-Task Learning in Natural Language Processing: An Overview}},
  year      = {2024},
  publisher = {Association for Computing Machinery},
  volume    = {56},
  number    = {12},
  journal   = {ACM Comput. Surv.}
}

@inproceedings{Liu2018,
  title     = {{Neural Multitask Learning for Simile Recognition}},
  author    = {Liu, Lizhen  and
               Hu, Xiao  and
               Song, Wei  and
               Fu, Ruiji  and
               Liu, Ting  and
               Hu, Guoping},
  booktitle = {Proceedings of the 2018 Conference on Empirical Methods in Natural Language Processing},
  year      = {2018},
  publisher = {Association for Computational Linguistics},
  pages     = {1543--1553}
}

@inproceedings{Grinsztajn2022,
  author    = {Grinsztajn, Leo and Oyallon, Edouard and Varoquaux, Gael},
  booktitle = {Advances in Neural Information Processing Systems},
  pages     = {507--520},
  publisher = {Curran Associates, Inc.},
  title     = {{Why do tree-based models still outperform deep learning on typical tabular data?}},
  volume    = {35},
  year      = {2022}
}

@article{Borisov2024,
  author  = {Borisov, Vadim and Leemann, Tobias and Seßler, Kathrin and Haug, Johannes and Pawelczyk, Martin and Kasneci, Gjergji},
  journal = {IEEE Transactions on Neural Networks and Learning Systems},
  title   = {{Deep Neural Networks and Tabular Data: A Survey}},
  year    = {2024},
  volume  = {35},
  number  = {6},
  pages   = {7499-7519}
}

@inproceedings{McElfresh2023,
  author    = {McElfresh, Duncan and Khandagale, Sujay and Valverde, Jonathan and Prasad C, Vishak and Ramakrishnan, Ganesh and Goldblum, Micah and White, Colin},
  booktitle = {Advances in Neural Information Processing Systems},
  pages     = {76336--76369},
  publisher = {Curran Associates, Inc.},
  title     = {{When Do Neural Nets Outperform Boosted Trees on Tabular Data?}},
  volume    = {36},
  year      = {2023}
}

@inproceedings{Zhang2019,
  author    = {Zhang, Ya-Lin and Li, Longfei},
  title     = {{Interpretable MTL from Heterogeneous Domains using Boosted Tree}},
  year      = {2019},
  publisher = {Association for Computing Machinery},
  booktitle = {Proceedings of the 28th ACM International Conference on Information and Knowledge Management},
  pages     = {2053-2056}
}

@article{Aakarsh2023,
  title   = {{Dropped Scheduled Task: Mitigating Negative Transfer in Multi-task Learning using Dynamic Task Dropping}},
  author  = {Aakarsh Malhotra and Mayank Vatsa and Richa Singh},
  journal = {Transactions on Machine Learning Research},
  year    = {2023}
}

@article{Aoki2022,
  author  = {Aoki, Raquel and Tung, Frederick and Oliveira, Gabriel L.},
  journal = {IEEE/ACM Transactions on Computational Biology and Bioinformatics},
  title   = {{Heterogeneous Multi-Task Learning With Expert Diversity}},
  year    = {2022},
  volume  = {19},
  number  = {6},
  pages   = {3093-3102}
}

@article{Jiang2020Boosting,
  author  = {Jiang, Jiahui and Wang, Ruimin and Wang, Mengdi and Gao, Kaifu and Nguyen, Duc Duy and Wei, Guo-Wei},
  title   = {{Boosting Tree-Assisted Multitask Deep Learning for Small Scientific Datasets}},
  journal = {Journal of Chemical Information and Modeling},
  year    = {2020},
  volume  = {60},
  number  = {3},
  pages   = {1235--1244},
}

\end{document}